\documentclass[preprint, 10pt, sort&compress]{elsarticle}

\usepackage[margin=2.5cm]{geometry}
\usepackage{setspace}
\usepackage{lmodern}
\usepackage{amsmath,amssymb}
\usepackage{amsfonts,amsthm}
\usepackage{lineno}
\usepackage{graphicx}
\usepackage{bm, bbm}
\usepackage{algorithm}
\usepackage{algorithmic}
\usepackage{physics}
\usepackage{color}
\usepackage{pxfonts}
\usepackage[footnotesize,bf]{caption}
\usepackage{subcaption}
\usepackage{mathtools}
\usepackage{hhline}
\usepackage[T1]{fontenc}

\usepackage{caption}
\usepackage{graphicx}
\usepackage{float}
\usepackage{subcaption}

\usepackage{placeins}

\newcommand{\diag}{\operatorname{diag}}
\newtheorem{theorem}{Theorem}[section]

\newtheorem{definition}{Definition}[section]
\newtheorem{assumption}{Assumption}[section]

\newtheorem{remark}{Remark}[section]

\numberwithin{equation}{section}
\numberwithin{figure}{section}
\numberwithin{table}{section}

\makeatletter
\def\ps@pprintTitle{%
	\let\@oddhead\@empty
	\let\@evenhead\@empty
	\let\@oddfoot\@empty
	\let\@evenfoot\@oddfoot
}
\makeatother

\graphicspath{{figures/}}
\usepackage{multirow}
\usepackage{ulem}
\usepackage{tikz}
\usepackage{color}
\usepackage[]{xcolor}

\begin{document}
\begin{frontmatter}
\title{Frequency-adaptive Multi-scale Deep Neural Networks}
 \author[a,b]{Jizu Huang}
  \author[a,b]{Rukang You}
   \author[a,b]{Tao Zhou}
\date{}
 \address[a]{LSEC, Academy of Mathematics and Systems Science, Chinese Academy of Sciences, Beijing 100190, PR China.}
 \address[b]{School of Mathematical Sciences, University of Chinese Academy of Sciences, Beijing 100190, PR China.}
\cortext[cor]{Corresponding Author}

\begin{abstract}
Multi-scale deep neural networks (MscaleDNNs) with downing-scaling mapping have demonstrated superiority over traditional DNNs in approximating target functions characterized by high frequency features. However, the performance of MscaleDNNs heavily depends on the parameters in the downing-scaling mapping, which limits their broader application. In this work, we establish a fitting error bound to explain why MscaleDNNs are advantageous for approximating high frequency functions. Building on this insight, we construct a hybrid feature embedding to enhance the accuracy and robustness of the downing-scaling mapping. 
To reduce the dependency of MscaleDNNs on parameters in the downing-scaling mapping, we propose frequency-adaptive MscaleDNNs, which adaptively adjust these parameters based on a posterior error estimate that captures the frequency information of the fitted functions. Numerical examples, including wave propagation and  the propagation of a localized solution of the schr$\ddot{\text{o}}$dinger equation with a smooth potential near the semi-classical limit, are presented. 
These examples demonstrate that the frequency-adaptive MscaleDNNs improve accuracy by two to three orders of magnitude compared to standard MscaleDNNs.


\end{abstract}

\end{frontmatter}

\section{Introduction}
In recent years, deep neural networks (DNNs) have been widely applied in various fields including computer vision, speech recognition, natural language processing, and scientific computing based on partial differential equations (PDEs) \cite{brown1990statistical, vaswani2017attention, devlin2018bert, brown2020language, han2017deep, yu2018deep, zang2020weak, tran2019dnn, han2018solving, han2017deep2, he2018relu, liao2019deep, raissi2019physics, strofer2019data, wang2020mesh}. However, several challenges remain in applying conventional DNNs to computational science and engineering problems. One significant challenge is the inherent limitation of DNNs in effectively handling data with high frequency content, as demonstrated by the Frequency Principle (F-Principle). While many DNNs can quickly learn the low frequency content of data with good generalization, they are inadequate  with high frequency data \cite{rahaman2019spectral, xu2019frequency, xu2019training, zhang2019explicitizing, xu2018understanding}. Extending DNNs to solving high frequency or multi-scale problems and establishing a framework to estimate fitting error are crucial issues.


The error bounds of DNNs when fitting low frequency data have been rigorously studied in recent theoretical work \cite{langer2021approximating, lu2021deep, shen2019deep}. These studies show that the fitting error of DNNs is primarily influenced by the intrinsic properties of the function and its $C^q$ norm. To ensure fast convergence and a small fitting error, the depth of the DNNs must exceed a sufficiently large integer, which depends on the $C^q$ norm of the function being fitted. For functions with high frequency, the $C^q$ norm contains high frequency information and tends to be large, necessitating a potentially greater depth for DNNs and  highlighting the challenges faced by DNNs when handling high frequency data. Establishing a fitting error estimate that only depends on low order norm of fitted function, such as the $C^0$ norm, will guide us in designing DNNs capable of solving high frequency or multi-scale problems. 


Leveraging insights derived from the F-Principle, the learning behaviors of NDDs are influenced by the frequency characteristics of the data or solutions. Incorporating frequency information into the design of neural network architectures can enhance the learning process. In computer vision, several studies have shown that accounting for image frequencies can significantly improve both generalization and training speed \cite{deng2018learning, pan2018learning, wu2020multigrid}. In these tasks, the frequency components within an input (i.e. an image) concerning spatial locations are often identifiable prior to learning. Recently, multi-scale DNNs (MscaleDNNs) 
have demonstrated a significant enhancement over standard DNNs in addressing high frequency and multi-scale problems \cite{biland2019frequency, cai2020phase, liu2020multi, xu2019training}. MscaleDNNs achieve this by converting high frequency content into lower frequency representations using radial down-scaling mapping. This radial mapping is independent of dimensionality, making MscaleDNNs well-suited for high-dimensional problems. Numerical experiments have shown that MscaleDNNs are effective for solving linear elliptic PDEs with high frequency components  \cite{liu2020multi}. Unlike the frequency in computer vision tasks, the frequency in MscaleDNNs refers to the response frequency of the mapping from input to output, which is problem-dependent and often unknown. When the response frequency is unavailable, there remains considerable room for improving the accuracy and efficiency of MscaleDNNs.

From an alternative perspective, Neural Tangent Kernel (NTK) theory has been used to model and understand the behavior of DNNs \cite{jacot2018neural}. After applying spectral decomposition on the NTK, the training error was decomposed into the NTK's eigen-spaces. During training, DNNs tend to first learn the objective function along the characteristic directions of the NTK with larger eigenvalues \cite{tancik2020fourier} and exhibit a slower convergence rate when learning high frequency functions, a phenomenon commonly known as "spectral bias" \cite{rahaman2019spectral}. As reported by \cite{zhong2019reconstructing, rahimi2007random}, a Fourier features mapping of input enables DNNs to efficiently fit higher frequency functions. When solving multi-scale PDEs, based on the NTK theory, a multi-scale random Fourier features mapping was employed in Physics-informed neural networks (PINNs) to enhance the robustness and accuracy of conventional PINNs \cite{wang2022and, wang2021eigenvector}.  Similarly, using NTK approach, \cite{wang2022spectral} also derived diffusion equation models in the spectral domain for the evolution of training errors of two-layer MscaleDNNs, designed to reduce the spectral bias of fully connected deep neural networks in approximating higher frequency functions.


This work aims to analyze and address the aforementioned shortcomings of DNNs in fitting high frequency data,  with a particular focus on designing effective DNNs for solving multi-scale PDEs. We begin by rigorously studying the fitting error of MscaleDNNs (DNNs with Fourier embedding) in fitting functions with high frequency, deriving a fitting error bound that is weakly dependent or independent of the high frequency of the fitted functions. This result highlights the advantages of MscaleDNNs and DNNs with random Fourier features over standard DNNs.
Inspired by this new theoretical analysis, we propose a hybrid feature embedding that combines the radial down-scaling mapping of MscaleDNNs with Fourier feature mapping, which offers improvements in both accuracy and robustness. 
To further enhance the accuracy and efficiency of MscaleDNNs, similar to the adaptive sampling of training points \cite{wu2023comprehensive, gao2023failure, tang2023pinns}, we develop a posterior error estimate to capture the frequency information of the fitted functions. Utilizing this frequency information, we introduce a Frequency-adaptive MscaleDNNs and extend this approach to solving multi-scale PDEs. A series of benchmarks are presented to demonstrate the accuracy and effectiveness of the proposed methods.

The remainder of this paper is organized as follows. In section \ref{pre}, we provide a brief overview of MscaleDNNs \cite{liu2020multi} and random Fourier feature \cite{wang2021eigenvector}. 
In section \ref{aff}, we analyze the fitting error of DNNs for multi-scale functions. Section \ref{fa} proposes the hybrid feature embedding, the posterior error estimate, and the Frequency-adaptive MscaleDNNs, respectively. In section \ref{nu}, we present a detailed evaluation of our proposed Frequency-adaptive MscaleDNNs using a variety of representative benchmark examples. Finally, Section \ref{co} concludes the paper.

\section{Preliminaries of MscaleDNNs and random Fourier features}
\label{pre}
DNNs fit a given function $f(\bm{x})$ by learning the parameters of 
networks through a process called training. During training, the DNNs iteratively adjust their
parameters (weights and bias) to minimize a predefined loss function 
$\mathcal{L}(\theta)$, which is given by:
\begin{equation*}
  \mathcal{L}(\theta)=\int_{\Omega}\left|f(\bm{x}) - f_\theta(\bm{x})\right|^2 \mathrm{d}\bm{x},
\end{equation*}
where $f_\theta(\bm{x})$ is the output of a neural network with parameters $\theta$. 
For a multi-scale function $f(\bm{x})$ that contains high frequency information,
the F-Principle \cite{rahaman2019spectral, xu2019frequency, xu2019training} 
demonstrates that $f_{\theta}(\bm{x})$ 
often encounters the "curse of high frequency”, making DNNs inefficient at learning the high frequency components of the multi-scale function $f(\bm{x})$.
This limitation reduces the accuracy of DNNs when solving PDEs with multi-scale solutions. To address this issue, a series of algorithms have been 
developed to overcome the high frequency 
curse inherent in general DNNs. In this section, we will briefly review two strategies aimed at improving the fitting of multi-scale functions.
The first approach is MscaleDNNs \cite{liu2020multi}, which incorporate the concept of radial 
scaling in the frequency domain. The second approach involves 
the use of Fourier feature networks \cite{wang2021eigenvector}, analyzed 
through the framework of NTK theory, 
for constructing network architectures that 
incorporate spatio-temporal and multi-scale random Fourier features.
\subsection{MscaleDNNs}
\label{ms}
MscaleDNNs \cite{liu2020multi} aim to reduce 
high frequency learning problems to low 
frequency learning problems by using a down-scaling mapping 
in phase space.
Let us consider a band-limited function $f(\bm{x})$ with $\bm{x}\in \mathbb{R}^d$, 
whose Fourier transform $\hat{f}(\bm{k}):={\cal F}[f(\bm{x})](\bm{k})$ has a compact support 
$\mathbb{K}\left(K_{\text{max}}\right) = \left\{\bm{k}\in \mathbb{R}^d, |\bm{k}|\leq K_{\text{max}}\right\}$. 
The compact support is then decomposed into the union of $W$ 
concentric annuli with uniform or nonuniform widths, e.g.,
\begin{equation*}
\mathbb{K}_i = \left\{\bm{k}\in \mathbb{R}^d, \ (i-1)K_0 \leq |\bm{k}|\leq i K_0\right\}, \ K_0=K_{\text{max}}/W, \ 1\leq i \leq W,
\end{equation*}
and 
\begin{equation*}
  \mathbb{K}(K_{\text{max}}) = \bigcup_{i=1}^W \mathbb{K}_i.
\end{equation*}
Based on this decomposition, $\hat{f}(\bm{k})$ can be rewritten as
\begin{equation*}
  \hat{f}(\bm{k}) = \sum_{i=1}^W \mathcal{X}_{\mathbb{K}_i}(\bm{k})\hat{f}(\bm{k})\triangleq \sum_{i=1}^W \hat{f}_i(\bm{k}),
\end{equation*}
where $\mathcal{X}_{\mathbb{K}_i}$ is the indicator function of the set $\mathbb{K}_i$ and 
\begin{equation*}
  \text{supp} \hat{f}_i(\bm{k}) \subset \mathbb{K}_i. 
\end{equation*}
Using the inverse Fourier transform, we obtain the corresponding decomposition in the physical 
space: 
\begin{equation*}
  f(\bm{x}) = \sum_{i=1}^W f_i (\bm{x}),
\end{equation*}
where 
\begin{equation*}
  f_i(\bm{x}) = \mathcal{F}^{-1}[\hat{f}_i(\bm{k})](\bm{x}).
\end{equation*}

\begin{figure}[htbp]
	\centering
	\includegraphics[width=0.4\linewidth]{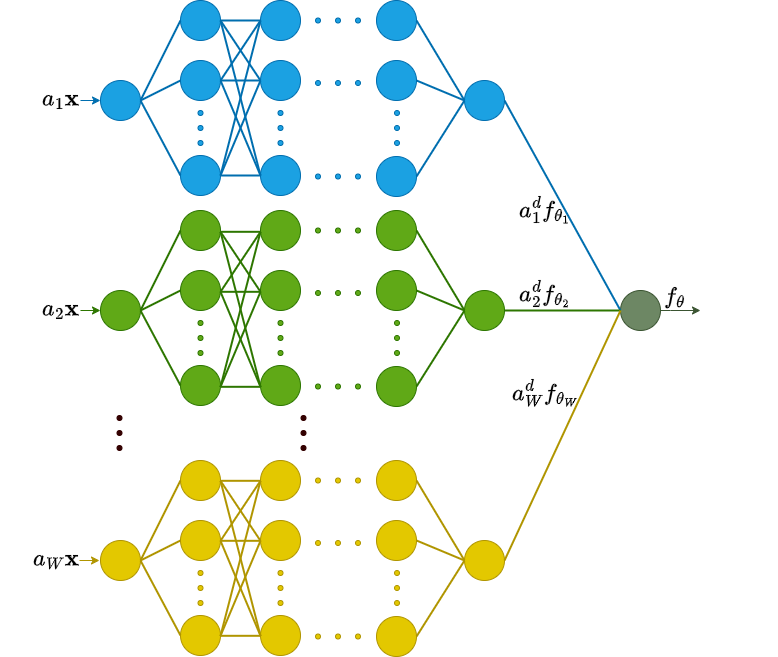}
	\caption{A structure of MscaleDNNs.}
	\label{msdnn0}
\end{figure}

MscaleDNNs introduce a down-scaling approach to convert 
the high frequency region $\mathbb{K}_i$ into a low frequency 
region by defining a scaled version of $\hat{f}_i(\bm{k})$ as 
\begin{equation*}
  \hat{f}_i^{(\text{scale})} (\bm{k}) = \hat{f}_i (a_i \bm{k}), \ a_i > 1.
\end{equation*}
The compact support of scaled version is
\begin{equation*}
  \text{supp}\hat{f}_i^{(\text{scale})}(\bm{k})\subset \left\{\bm{k} \in \mathbb{R}^d, 
  \frac{(i-1)K_0}{a_i}\leq |\bm{k}| \leq \frac{iK_0}{a_i}\right\}.
\end{equation*}
In the physical space, the down-scaling is denoted as
\begin{equation*}
  f_i^{(\text{scale})}(\bm{x}) = f_i \left(\frac{1}{a_i} \bm{x}\right) \frac{1}{a_i^d} \quad \text{or} \quad f_i(\bm{x})= a_i^df_i^{(\text{scale})}(a_i \bm{x}).
\end{equation*}
In MscaleDNNs, $a_i$ is chosen to be sufficiently large such that $f_i^{(\text{scale})}(\bm{x})$
possesses a low frequency spectrum. 
Typically, the value of $a_i$ is set to $2^{i-1}$.
According to the F-Principle
in conventional Deep Neural Networks \cite{xu2019frequency}, 
MscaleDNNs can facilitate rapid learning for $f_i^{(\text{scale})}(\bm{x})$, 
denoted as $f_{\theta_i}$ with $\theta_i$ being the parameters of MscaleDNNs.
Finally, MscaleDNNs approximate $f(\bm{x})$ as: 
\begin{equation}
  f_\theta(\bm{x}) = \sum_{i=1}^W a_i^d f_{\theta_i}(a_i \bm{x}).
  \label{eq21_1}
\end{equation}
MscaleDNNs transform the original input data $\bm{x}$ into $\{a_1\bm{x}, \dots, a_W\bm{x}\}$ and 
provide an efficient ansatz for approximating the function $f(\bm{x})$ with high frequency.
A typical network structure of MscaleDNNs is 
illustrated in Figure \ref{msdnn0}.

In summary, MscaleDNNs utilize a frequency domain scaling technique 
to develop a multi-scale capability for approximating target 
functions with rich frequency content. Specifically, 
several numerical simulations reported in \cite{liu2020multi} validate that MscaleDNNs are
an efficient, mesh-less, and easily 
implementable method for solving multi-scale PDEs. 
However, the performance of MscaleDNNs is sensitive to the choice of the scaling parameters $a_i$, which 
potentially limits their applicability. Additionally, MscaleDNNs increase
the width of neural networks by transforming the input $\bm{x}$ into $a_1\bm{x},\cdots, a_W\bm{x}$, leading to higher 
training costs due to the additional parameters. 
For example, the optimal parameters $a_i$ for the 
target function $f(x)=\sin(\pi x)+\sin(100\pi x)$ are ideally $a_1=\pi$ and 
$a_2 = 100\pi$. However, constructing an algorithm to find these optimal parameters without 
prior knowledge of the frequency distribution remains a significant challenge.

\subsection{DNNs with Fourier features}
\label{ra}
Recently, NTK theory has been employed to model and understand the behavior of DNNs \cite{jacot2018neural}. 
By introducing a semi-positive definite NTK and performing its spectral decomposition, 
it is possible to break down the training error into the NTK's eigen-spaces. 
During training, DNNs tend to first learn the objective function along the characteristic directions of the NTK with larger eigenvalues and gradually progress to learning components associated with smaller eigenvalues \cite{tancik2020fourier}. 
As a result, DNNs exhibit a slower convergence rate when learning high frequency functions, a phenomenon commonly known as "spectral bias" \cite{rahaman2019spectral}. Numerical 
simulations reported in \cite{zhong2019reconstructing} 
demonstrated that a heuristic sinusoidal mapping of input $x$ enables DNNs to 
efficiently fit higher frequency functions. This heuristic mapping is typically a special case of Fourier features \cite{rahimi2007random}.
A random Fourier feature mapping $\beta:\mathbb{R}^d \to \mathbb{R}^{2m}$ is 
generally defined as \cite{tancik2020fourier}:
\begin{equation}
  \beta[\bm{A}](\bm{x}) = \begin{bmatrix}
    \sin(\bm{A}\bm{x})\\
    \cos(\bm{A}\bm{x})
  \end{bmatrix},
  \label{fe2}
\end{equation}
where the entries of $\bm{A} \in \mathbb{R}^{m\times d}$ are sampled from a Gaussian 
distribution $\mathcal{N}(0, \sigma^2)$ with $\sigma > 0$ being a user-specified 
hyper-parameter. By employing random Fourier feature mapping, the NTK of DNNs achieves
stationarity (shift-invariance), which acts as a convolution kernel over the input domain \cite{tancik2020fourier}.
This mapping also allows for the regulation of the NTK's bandwidth, resulting in improved training speed and generalization \cite{tancik2020fourier}. 

As shown in \cite{wang2021eigenvector}, the 
frequency of eigen-functions for the NTK is influenced by entries of $\bm{A}$, which are 
sampled from Gaussian distribution $\mathcal{N}(0, \sigma^2)$. Choosing a
larger value for $\sigma$ increases the likelihood of capturing high frequencies. Therefore, 
a mapping with large $\sigma$ can help alleviate the challenges posed by spectral bias, allowing
networks to more effectively learn functions with high frequency. However, random Fourier
feature mappings initialized with a large $\sigma$ may cause over-fitting and do not always 
benefit the DNNs, especially for target functions with low frequencies. In order to fit functions with 
multi-scale frequencies, multiple random Fourier feature mappings with different $\sigma_i, \ i=1, \cdots, \ N_{\sigma}$ were
proposed in \cite{wang2021eigenvector}. This new network architecture is named DNNs with Fourier features.
Similar to the MscaleDNNs, the choice of $\sigma_i$ is problem-dependent. \par

In summary, both MscaleDNNs and DNNs with Fourier features can effectively fit multi-scale functions 
by incorporating frequency scaling 
mappings into DNNs. However, the performance of these DNNs is heavily dependent on parameter selection, which potentially
limits their applicability. In the following section, we will develop a theoretical framework 
to better understand the performance of DNNs with frequency scaling mappings. This framework will further inspire the development of frequency-adaptive MscaleDNNs in section \ref{fa}.

\section{Error estimate of DNNs in fitting functions with high frequency}
\label{aff}
We establish a framework to estimate fitting error for functions with high frequency components, demonstrating that the fitting error of DNNs is reduced by incorporating appropriate down-scaling mappings. To proceed, we first introduce some definitions and notations that will be used throughout the remainder of this paper. {The sets of natural numbers,
natural numbers including 0, and real numbers are denoted by $\mathbb{N}, \ 
\mathbb{N}_0$, and $\mathbb{R}$, respectively. Let $\bm{j}=(j_1,\ldots, \ j_d)^T \in \mathbb{N}_0^d$ be a $d$-dimensional vector, referred to as a multi-index.}
\begin{definition}
  Assume $p=q+s$ with $q\in \mathbb{N}_0$ and $0<s\leq 1$. A function $f:\mathbb{R}^d \to \mathbb{R}$ is called 
  $(p, C_0)$-smooth, if for every $\bm{j}=(j_1, \dots, j_d)^T\in \mathbb{N}_0^d$ with $|\bm{j}|_1=q$, the 
  partial derivative $\partial^q f /(\partial x_1^{j_1} \dots \partial x_d^{j_d})$ exists and satisfies 
  \begin{equation*}
    \left|\frac{\partial^q f(\bm{x})}{\partial x_1^{j_1} \dots \partial x_d^{j_d}} - \frac{\partial^q f(\bm{y})}{\partial x_1^{j_1} \dots \partial x_d^{j_d}}\right| \leq C_0\|\bm{x}-\bm{y}\|^s 
  \end{equation*}
  for all $\bm{x},\bm{y}\in \mathbb{R}^d$, where $\|\cdot\|$ denotes the Euclidean norm and $|\bm{j}|_1 = \sum_{i=1}^d|j_i|$.
  \label{defi1}
\end{definition}

\begin{definition}
  Assume $f_{\bm{\theta}, L, \iota}$ is the output of standard neural network with sigmoid activation function $\text{act}(x)=\frac{1}{1+\exp(-x)}$,
  depth L, width $\iota$, and parameters $\bm{\theta}$. We define a network class as 
  \begin{equation*}
    \mathcal{F}(L,\iota, \delta )=\left\{f:= f_{\bm{\theta}, L, \iota} \ \text{with} \ \|\bm{\theta}\|_{\infty} \leq \delta\right\}.
  \end{equation*}
  \label{defi2}
\end{definition}
\begin{definition}
  Let us denote $C' y\leq x\leq C'' y$ as $x\lesssim y$, where $C'$ and $ C''$ are two positive constants.  
  \label{defi3}
\end{definition}
\begin{definition}
  For a given feature mapping $\Phi(\bm{x}):=\bigl(\Phi_1(\bm{x}),\ \Phi_2(\bm{x}),\ \dots,\ \Phi_N(\bm{x})\bigr)^T$ that maps $\mathbb{R}^d \to \mathbb{R}^{ND}$ with 
  $\Phi_i:\mathbb{R}^d \to \mathbb{R}^D$, we define a feature mapping network class as 
  \begin{equation*}
    \widetilde{\mathcal{F}}(N, L,\iota, \delta )=\left\{f:= WG + b \ \text{with} \ G=\left(g_1\left(\Phi_1(\bm{x})\right),\dots,g_N\left(\Phi_N(\bm{x})\right)\right)^T \, \text{and} \  g_i \in \mathcal{F}(L,\iota, \delta)\right\}, 
  \end{equation*}
  where $W \in \mathbb{R}^{1\times N}$ and $ b\in \mathbb{R}$ are optimizable parameters of neural networks. 
  \label{defi4}
\end{definition}

With Definition \ref{defi4}, the output of 
MscaleDNNs belongs to a  feature mapping neural network class $\widetilde{\mathcal{F}}(N, L, \iota, \delta)$ 
with $\Phi(\bm{x}) = ( \bm{x}, \dots, 2^{N-1} \bm{x})^T$. 
Similarly, for DNNs with Fourier features as described in \cite{rahimi2007random},
the output $f_{\text{net}} \in \widetilde{\mathcal{F}}(N, L, \iota, \delta)$
with $\Phi(\bm{x}) = \beta[\bm{A}](\bm{x})$ as defined by (\ref{fe2}). 
We introduce the following norms for a smoothing function $f$
\begin{equation*}
  \|f\|_{L^\infty([-h,h]^d)}:= \text{max}\left\{|f(\bm{x})|: \bm{x}\in [-h,h]^d\right\} \quad 
  \text{for any} \quad f\in C\left([-h,h]^d\right),
\end{equation*}
\begin{equation*}
  \|f\|_{C^q([-h,h]^d)}:= \text{max}\left\{\|\partial^{\bm{j}}f\|_{L^{\infty}([-h,h]^d)}:|\bm{j}|_1\leq q, \bm{j}\in \mathbb{N}^d\right\} \quad 
  \text{for any} \quad f\in C^q\left([-h,h]^d\right),
\end{equation*}
and
\begin{equation*}
  \|f\|_{C^{q, 1}([-h,h]^d)}:= \text{max}\left\{\|\partial^{\bm{j}}f\|_{L^{\infty}([-h,h]^d)}: 1 \leq |\bm{j}|_1\leq q, \bm{j}\in \mathbb{N}^d\right\} \quad 
  \text{for any} \quad f\in M_q\left([-h,h]^d\right),
\end{equation*}
where $M_q\left([-h,h]^d\right) = C^q\left([-h,h]^d\right)/\mathbb{R}$.
Several existing works reveal the distance between a neural network function and a given smooth function for standard DNNs \cite{langer2021approximating, lu2021deep, shen2019deep}.  
We introduce one of these results in the following theorem, which will 
be used to construct a theoretical analysis result
for functions with high frequency.
\begin{theorem}
  Assume $1\leq h<\infty$, $p=q+s$ with $q\in \mathbb{N}_0$, and $s\in (0,1]$. Let 
  $f:\mathbb{R}^d \to \mathbb{R}$ be a $(p, C_0)$-smooth function satisfying
  \begin{equation*}
    \|f\|_{C^q([-2h,2h]^d)}\leq C_1
  \end{equation*}
  for constants $C_0\geq1$ and $C_1>0$. Let $\text{act}(x)$ be the sigmoid activation function.
  For any $M\in \mathbb{N}$ sufficiently large, there exists a neural network $f_{\text{net}}$ in the network class $\mathcal{F}(L,\iota,\delta)$ such that
  \begin{equation*}
    \|f_{\text{net}} - f\|_{\infty,[-h, h]^d}\leq \frac{C_2 \left({\max}\left\{h, \|f\|_{C^{q,1}\left([-h,h]^d\right)}, \|f\|_{L^\infty\left([-h,h]^d\right)} \right\}\right)^{5q + 3}}{M^{2p}}
  \end{equation*}
  holds for a constant $C_2>0$. Here
  $M^{2p}\geq {\max}\left\{2c_2\left(\max\left\{ h,\|f\|_{C^{q,1}\left([-h,h]^d\right)}, \|f\|_{L^\infty\left([-h,h]^d\right)}\right\}\right)^{5q+3}, c_3, 2^d, 12d \right\}$ with $c_2$ and $c_3$ being constants, 
  $L = C_4(d,q), \ \iota = C_5(d, q, M),\ \text{and} \ \delta =C_6\left(d,M,h,{\max}\left\{h, \|f\|_{C^{q,1}\left([-h,h]^d\right)}, \|f\|_{L^\infty\left([-h,h]^d\right)}\right\}\right).$
  \label{thm0}
\end{theorem}
\begin{remark}
  In Theorem \ref{thm0}, the constants $C_4$, $C_5$, and $C_6$ are defined as:
  \begin{equation*}
    \begin{split}
      L&=C_4(d,q)=8+\left\lceil \log_2\left({\max}\{d,q+1\}\right) \right\rceil, \\
      \iota&=C_5(d,q,M)=2^d\left({\max}\left\{\left(C^{d}_{d+q}
      +d \right)M^d (2+2d)+d,\, 4(q+1) C^{d}_{d+q} \right\}
      +M^d(2d + 2) + 12d \right), \\
      \delta&=C_6\left(d,M,h,{\max}\left\{h, \|f\|_{C^{q,1}([-h,h]^d)}, \|f\|_{L^\infty([-h,h]^d)}\right\}\right)\\
      &=c_6 \exp{6\times 2^{2(d+1)+1}dh} M^{10p+2d+10}\left({\max}\left\{h, \|f\|_{C^{q,1}([-h,h]^d)}, \|f\|_{\infty, ([-h,h]^d)} \right\}\right)^{12},
    \end{split}
  \end{equation*}
  where $c_6$ is a positive constant.
  In the following part of this paper, we always denote $C_4,\ C_5$ and $   C_6$ as functions
  defined by the above equations.
\end{remark}

According to Theorem \ref{thm0}, the fitting error of a standard DNNs depends on $M$, 
whose lower bound is a function of  ${\max}\left\{2c_2\left(\max\left\{ h,\|f\|_{C^{q,1}([-h,h]^d)}, \|f\|_{L^\infty([-h,h]^d)}\right\}\right)^{5q+3}, c_3, 2^d, 12d \right\}$. For a function $f$ 
containing high frequency, the norm $\|f\|_{C^{q,1}([-h,h]^d)}$ 
is typically much large, resulting a relatively large lower bound for $M$ and $\max\left\{ h,\|f\|_{C^{q,1}([-h,h]^d)}, \|f\|_{L^\infty([-h,h]^d)}\right\}=\|f\|_{C^{q,1}([-h,h]^d)} $. 
By setting $M=\|f\|^{\gamma}_{C^{q,1}([-h,h]^d)}$, the error bound in Theorem \ref{thm0} 
can be expressed as $\|f_{net}-f\|_{\infty,[-h,h]^d}\leq C_2\left( \|f\|_{C^{q,1}([-h,h]^d)}\right)^{5q+3-2p\gamma}$. To ensure small fitting errors, the parameter $\gamma $ should be larger than $\frac{5q+3}{2p}$ and $M$ should be also larger than $\|f\|^{\frac{5q+3}{2p}}_{C^{q,1}([-h,h]^d)}$, resulting in a standard DNNs with a relatively large width 
$\iota=C_5(q,d,M)$. Furthermore, since the parameters $\bm{\theta}$ must satisfy  $\|\bm{\theta}\|_{\infty}\leq\delta$, the feasible set of parameters $\bm{\theta}$ also becomes relatively large, which leads to high training costs for high frequency functions $f$. This explains why the standard DNNs suffer from "curse of high frequency".
Our next goal is 
to estimate the fitting error of MscaleDNNs for a high frequency 
function $f$ that meets the following assumption.
\begin{assumption}
  \label{ass1}
  Assume that $f(\bm{x})\in C^q[-h,h]^d$ with $q\geq 1$ is a high frequency function satisfying 
  \begin{equation*}
    \|\partial^{\bm{j}} f\|_{L^{\infty}([-2h,2h]^d)} \lesssim k^{|{\bm{j}}|_1}  \quad \text{for} \quad 1\leq |{\bm{j}}|_1 \leq q, \quad\text{ and }\quad \|f\|_{L^{\infty}([-2h,2h]^d)}\leq C.
  \end{equation*}
  Here $k$ is a positive constant determined by the high frequency 
  and $C$ is a constant independent of $k$.
\end{assumption}\par 

The following Theorem demonstrates how the 
down-scaling mapping $\Phi(\bm{x})=k \bm{x}$ in the MscaleDNNs reduces 
the fitting error of neural networks.
\begin{theorem}
  \label{thm1}
  Assume $1\leq h<\infty$, $p=q+s$ with $q\in \mathbb{N}_0$, and $s\in (0,1]$. 
  For $C_0\geq 1$, let $f:\mathbb{R}^d \to \mathbb{R}$ be a $(p, C_0)$-smooth function satisfying assumption \ref{ass1}. 
  For any $M\in \mathbb{N}$ sufficiently large, there exists a neural network $f_{\text{net}}$ in the 
  network class $\widetilde{\mathcal{F}}(1,L,\iota,\delta)$ with $\Phi(\bm{x}) = k \bm{x}$ such that 
  \begin{equation*}
    \|f_{\text{net}} - f\|_{\infty,[-h, h]^d} \leq \frac{C_2 \left({\max}\left\{k h, C_1\right\}\right)^{5q + 3}}{M^{2p}}
  \end{equation*}
  holds for constants $C_1>0$ and $C_2>0$ independent of $k$. Here 
  $M^{2p}\geq {\max}\big\{2c_2\left(\max\{ kh,C_1\}\right)^{5q+3}, c_3, 2^d, 12d \big\}$, 
  $L = C_4(d,q), \ \iota = C_5(d, q, M), \ \text{and} \ \delta = C_6\left(d,M,h,{\max}\left\{k h, C_1 \right\}\right).$
\end{theorem}
\begin{proof}
  Let $F(\bm{y}) = f(\bm{y}/k) = f(\bm{x})$ with $\bm{y}=k \bm{x}$ and $k > 0$. 
  By the chain rule for derivation, we have
  \begin{equation*}
    \partial^{\bm{j}}_{\bm{x}} f(\bm{x}) = \partial^{\bm{j}}_{\bm{x}} (F(k\bm{x})) = k^{|\bm{j}|_1}\partial^{\bm{j}}_{\bm{y}}F(\bm{y}),
  \end{equation*}
  and 
  \begin{equation*}
    \|\partial^{\bm{j}}_{\bm{x}} f\|_{L^\infty([-h,h]^d)} = k^{|\bm{j}|_1}\cdot \|\partial^{\bm{j}}_{\bm{y}}F\|_{L^\infty([-k h, k h]^d)}.
  \end{equation*}
  According to Assumption \ref{ass1}, we obtain 
  \begin{equation*}
    \|F\|_{L^\infty([-2k h, 2k h]^d)} = \|f\|_{L^\infty([-2h, 2h]^d)} \leq C \quad \text{and} \quad \|F\|_{C^{q,1}([-2k h, 2k h]^d)} \leq C',
  \end{equation*}
  where $C$ and $C'$ are two constants independent of $k$. \par 
  According to Theorem \ref{thm0}, there exists a neural network $F_{\text{net}}$ in $\mathcal{F}(L, \iota, \delta)$ such that
  \begin{equation*}
    \begin{split}
      \|F_{\text{net}} - F\|_{\infty,[-k h, k h]^d} & \leq \frac{C_2 \bigg(\text{max}\big\{k h,  \|F\|_{C^{q,1}([-k h,k h]^d)}, \|F\|_{\infty, ([-k h,k h]^d)}\big\}\bigg)^{5q + 3}}{M^{2p}} \\
      & \leq \frac{C_2 \bigg(\text{max}\big\{k h, C_1 \big\}\bigg)^{5q + 3}}{M^{2p}},
    \end{split}
  \end{equation*}
  where $C_1=\max\{C, C'\}$, $C_2$ are constant independent of $k$. Here, $M^{2p}\geq {\max}\big\{2c_2\left(\max\{ kh,C_1\}\right)^{5q+3}, c_3, 2^d, 12d \big\}$, 
  $L = C_4(d,q), \ \iota = C_5(d, q, M), \ \text{and} \ \delta = C_6\left(d,M,kh,\text{max}\left\{k h, C_1 \right\}\right)$.
  By taking $f_\text{net}(\bm{x}) = W F_{\text{net}}(\Phi(\bm{x})) + b$ with $W=1$ and $b=0$, it follows
  \begin{equation*}
    \|f_{\text{net}} - f\|_{\infty,[-h, h]^d} \leq \frac{C_2 \bigg(\text{max}\big\{k h, C_1 \big\}\bigg)^{5q + 3}}{M^{2p}},
  \end{equation*}
  which completes the proof of Theorem \ref{thm1}.
\end{proof}

For a function $f$ 
containing high frequency, 
$kh$ is typically larger than $  C_1,\ C_3,\ 2^d$, and $12d$. 
By denoting $M=(kh)^\gamma$, the error bound in Theorem \ref{thm1} 
can be expressed as $\|f_{net}-f\|_{\infty,[-h,h]^d}\leq C_2\left( kh\right)^{5q+3-2p\gamma}$. To achieve small fitting errors, the parameter $M$ should be larger than $(kh)^{\frac{5q+3}{2p}}$. According to Assumption \ref{ass1}, we have $\|f\|_{C^{q,1}([-h,h]^d)}\geq C k^{q} \gg hk$ for $q\geq 2$ and a large $k$. In this situation, the width of MscaleDNNs is smaller than that of standard DNNs. Additionally, the feasible set of parameters $\bm{\theta}$ also becomes relatively smaller. 
This is why MscaleDNNs are both efficient and accurate in fitting functions 
with high frequency. 

Next, we estimate the fitting error
for DNNs with Fourier features. Suppose 
$f(\bm{x})=F(\sin(\bm{k} \cdot \bm{x}))$ or $f(\bm{x})=F(\cos(\bm{k}\cdot \bm{x}))$ with $\bm{k}=\{k_1, \dots, k_d\}$, 
where $F$ is a smooth function with 
its $C_{q}$ norm bounded by a constant independent of $\bm{k}$.
For DNNs with Fourier feature $\beta[\diag (\bm{k})](\bm{x})$, the 
following theorem shows that the fitting error for $f$ is independent 
of frequency $\bm{k}$. \par
\begin{theorem}
  Assume $f(\bm{x})=F(\sin(\bm{k}\cdot\bm{x}))$ or $f(\bm{x})=F(\sin(\bm{k}\cdot\bm{x}))$ 
  with $F:\,[-2,2]\rightarrow \mathbb{R}  $ being a $(p,C_0)$-smooth function and satisfying {$\|F\|_{C^q([-2,2])} \leq C_1$}. Here
  $\bm{x}\in[-\pi, \pi]^d $ and $C_0,C_1$  are constants independent of $\bm{k}$.
  If $L = C_4(1,q), \ \iota = C_5(1, q, M), \ \text{and} \ \delta = C_6(1,M,1,\max\{ 1,C_1\})$, there 
  exists a neural network $f_{\text{net}}$ in the 
  network class $\widetilde{\mathcal{F}}(1, L,\iota,\delta)$ with $\Phi(\bm{x}) = \sin(\bm{k}\cdot  \bm{x})$ or $\Phi(\bm{x}) = \cos(\bm{k} \cdot \bm{x})$, 
  such that
  \begin{equation*}
    \|f_{\text{net}} - f\|_{\infty,[-\pi, \pi]^d} \leq \frac{C_2'}{M^{2p}},
  \end{equation*}
  where $M^{2p}\geq {\max}\left\{2c_2\left(\max\{ 1,C_1\}\right)^{5q+3}, c_3,  12 \right\}$ and $C_2'>0$ is a constant independent 
of frequency $\bm{k}$.
  \label{thm_co1}
\end{theorem}
\begin{proof}
  We only show the proof for $f(\bm{x})=F(\sin(\bm{k} \cdot \bm{x}))$. 
  The proof for $f(\bm{x})=F(\cos(\bm{k} \cdot \bm{x}))$ follows in a similar manner.
  By setting $y=\sin(\bm{k} \cdot \bm{x})$, {we have 
  $F(y)$ being a $(p,C_0)$-smooth function and satisfying
  \begin{equation*}
    \|F\|_{C^q([-2, 2])} \leq C_1,
  \end{equation*}
  }where $C_1$ is a constant independent of $\bm{k}$.
  Following the proof of Theorem\ref{thm0}, there exists a neural 
  network $F_{\text{net}}$ in $\mathcal{F}(1,L,\iota,\delta)$ such that
  \begin{equation*}
    \begin{split}
      \|F_{\text{net}} - F\|_{\infty,[-1, 1]} & \leq \frac{C_2 \bigg({\max}\big\{1,  \|F\|_{C^{q,1}([-1,1])}, \|F\|_{\infty, ([-1,1])}\big\}\bigg)^{5q + 3}}{M^{2p}} \\
      & \leq \frac{C_2 \left({\max}\big\{1,\,C_1\big\}\right)^{5q + 3}}{M^{2p}}=\frac{C'_2}{M^{2p}},
    \end{split}
  \end{equation*}
  where $C_2 $ and $C'_2 $  are constants independent of $\bm{k}$, 
  $M^{2p}\geq {\max}\big\{2c_2\left(\max\{ 1,C_1\}\right)^{5q+3}, c_3,  12 \big\}$, 
  $L = C_4(1,q), \ \iota = C_5(1, q, M), \ \text{and} \ \delta = C_6(1, M,1,{\max}\{1,\,C_1\})$.
  By taking $f_\text{net}(\bm{x}) = W F_{\text{net}}(\Phi(\bm{x})) + b$ with $W=1, \ b=0$, and $\Phi(x)=\sin(\bm{k}\cdot \bm{x})$, we obtain 
  \begin{equation*}
    \|f_{\text{net}} - f\|_{\infty,[-\pi, \pi]^d}=\|F_{\text{net}}-F\|_{\infty,[-1,1]} \leq \frac{C'_2}{M^{2p}},
  \end{equation*}
  which completes the proof of Theorem \ref{thm_co1}.
\end{proof}

Based on the above results, the following theorem establish a frequency independent fitting error 
estimate for a band-limited function $f$, which is the 
main theoretical result of this paper.
\begin{theorem}
  Assume $f\in L^2([-\pi, \pi]^d)$ is a band-limited function, which can be extended as
  \begin{equation*}
    f(\bm{x})  = \sum_{\bm{k}\in \mathbb{B}} [b_{\bm{k}}\cos(\bm{k} \cdot \bm{x}) +  c_{\bm{k}} \sin(\bm{k}\cdot \bm{x})],
  \end{equation*}
  where $b_{\bm{k}}$ and $c_{\bm{k}}$ with $\bm{k}\in \mathbb{B}$ are bounded constants.
  Here $\mathbb{B}$ is the compact support of Fourier transform $\hat{f}$ with $|\mathbb{B}|=N$. 
  For Fourier features $\Phi(\bm{x})=(\ldots, \sin(\bm{k}\cdot \bm{x}), \cos(\bm{k}\cdot \bm{x}),\ldots)$ with 
  $\bm{k}\in \mathbb{B}$, if $L = C_4(1,q), \ \iota = C_5(1, q, M), \ \text{and} \ \delta = C_6(1,M,1,1)$,  
  there exists a neural network $f_{\text{net}}$ in the 
  network class $\widetilde{\mathcal{F}}(2N, L,\iota,\delta)$, such that 
  \begin{equation*}
    \|f_{\text{net}} - f\|_{\infty,[-\pi, \pi]^d} \leq \frac{C_2\sqrt{N}}{M^{2p}},
  \end{equation*}
  holds for constant $C_2>0$. Here $M^{2p}\geq {\max}\big\{2c_2, c_3,  12 \big\}$.
  \label{thm2}
\end{theorem}
\begin{proof}
  Based on the assumptions in Theorem \ref{thm2}, we have:
  \begin{equation*}
    f(\bm{x})=\sum_{\bm{k}\in \mathbb{B}}[b_{\bm{k}}\cos(\bm{k}\cdot\bm{x})+ c_{\bm{k}}\sin(\bm{k}\cdot\bm{x})].
  \end{equation*}
  Since $F(y)=y$ is a $(p,1)$-smooth function for any $p\geq1$ and satisfies $ \|F\|_{C^q([-1,1])}\leq 1$, according to Theorem \ref{thm_co1}, there exist $\tilde{f}^1_{\text{net},\bm{k}}$ and $\tilde{f}^2_{\text{net},\bm{k}}$ in the network class $\widetilde{\mathcal{F}}(1, L,\iota,\delta)$, such that: 
  \begin{equation*}
    \|\tilde{f}^1_{\text{net},\bm{k}} - \sin(\bm{k}\cdot\bm{x})\|_{\infty,[-\pi, \pi]^d} \leq \frac{C'_2}{M^{2p}} \ \text{~and~} \ \|\tilde{f}^2_{\text{net},\bm{k}} - \cos(\bm{k}\cdot\bm{x})\|_{\infty,[-\pi, \pi]^d} \leq \frac{C'_2}{M^{2p}}
  \end{equation*}
  hold for $\bm{k}\in\mathbb{B}$ and $M^{2p}\geq {\max}\big\{2c_2, c_3,  12 \big\}$. Here $L = C_4(1,q), \ \iota = C_5(1, q, M), \ \text{and} \ \delta = C_6(1, M,1,1)$.
  By taking $f_{\text{net}}(\bm{x}) = \sum_{\bm{k}\in \mathbb{B}}(b_{\bm{k}}\tilde{f}^1_{\text{net},\bm{k}}+c_{\bm{k}}\tilde{f}^2_{\text{net},\bm{k}})$,
  we have $f_{\text{net}}\in \widetilde{\mathcal{F}}(2N, L,\iota,\delta)$. The distance between $f$ and 
  $f_{\text{net}}$ is given by: 
  \begin{equation}
    \label{proof1}
    \begin{split}
      \|f_{\text{net}} - f\|_{\infty,[-\pi, \pi]^d} & \leq \sum_{\bm{k}\in\mathbb{B}}\left( |b_{\bm{k}}| \|\tilde{f}^1_{\text{net},\bm{k}}(\bm{x}) - \sin(\bm{k}\cdot \bm{x})\|_{\infty,[-\pi, \pi]^d}+|c_{\bm{k}}|\|\tilde{f}^2_{\text{net},\bm{k}}(\bm{x}) - \cos(\bm{k}\cdot\bm{x})\|_{\infty,[-\pi, \pi]^d}\right)\\
      & \leq  \frac{C'_2}{M^{2p}}\sum_{\bm{k}\in \mathbb{B}}\left( {|b_{\bm{k}}|} + {|c_{\bm{k}}|}\right).
    \end{split}
  \end{equation}
  For $f(\bm{x})\in L^2([-\pi, \pi]^d)$, it follows from Parseval's theorem that: 
  \begin{equation}
    \label{proof2}
    \begin{split}
      \sum_{\bm{k}\in \mathbb{B}} \left(|b_{\bm{k}}| + |c_{\bm{k}}|\right) &\leq \sqrt{{2N}\left(\sum_{\bm{k}\in \mathbb{B}}\left( |b_{\bm{k}}|^2 + |c_{\bm{k}}|^2\right)\right)} \\
      & \leq \frac{\sqrt{2N}} {{(2\pi)^{d/2}}}\|f\|_{L^2([-\pi,\pi]^d)}  \leq C \sqrt{N} ,
    \end{split} 
  \end{equation}
  where $C$ is constant determined by $\|f\|_{L^2([-\pi,\pi]^d)}$.
  By combining \ref{proof1} and \ref{proof2}, we obtain
  \begin{equation*}
    \|f_{\text{net}} - f\|_{\infty,[-\pi, \pi]^d} \leq \frac{C_2\sqrt{N}}{M^{2p}},
  \end{equation*}
  where $C_2$ is a constant and $M^{2p}\geq {\max}\big\{2c_2, c_3,  12 \big\}$.  The proof of Theorem \ref{thm2} is complete.
\end{proof}

According to Theorem \ref{thm2}, if the frequency compact support $\mathbb{B}$ of the 
fitted function $f$ is known, the fitting error of the DNNs with multiple 
Fourier features depends on $|\mathbb{B}|=N$ but independent of the specific frequency $\bm{k}$. 
This suggests that DNNs with Fourier features have an advantage when fitting functions with known high frequencies.   However, in many applications, the frequency compact support of the 
fitted function is often unknown. In the cases, when using MscaleDNNs and DNNs with 
Fourier features, a sufficiently large or randomly chosen frequency set is typically employed to ensure that the frequency compact support of the function is approximately captured. This, however, increases the width of the neural networks, leading to higher training costs due to the additional parameters.  Therefore, estimating the frequency compact support of the fitted function is crucial for accelerating the convergence of the neural network. To efficiently determine the frequency compact support is a key motivation behind the development of frequency-adaptive MscaleDNNs.

\section{Hybrid feature embedding and frequency-adaptive MscaleDNNs}
\label{fa}
In subSection \ref{fe}, we begin by examining the accuracy and robustness of the different down-scaling mappings discussed in Section \ref{pre}, which motivates us to propose a hybrid feature embedding. Next, we establish a posterior error estimates for high frequency functions, aiding in capturing the frequency information of the fitted functions. Finally, we propose a frequency-adaptive MscaleDNN to solve multi-scale PDEs in subsection \ref{fr}.

\subsection{Hybrid feature embedding}
\label{fe}
\begin{figure}[htbp]
	\centering
	\includegraphics[width=0.8\linewidth]{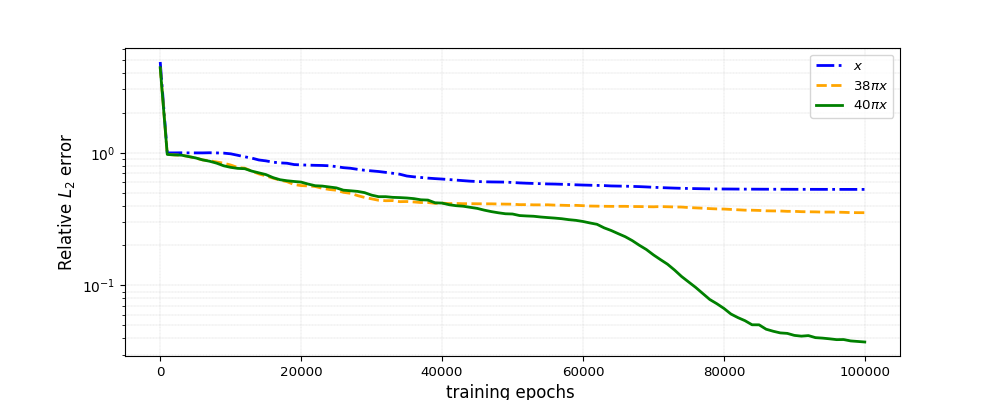}
	\caption{The relative $L_2$ error at each training epoch: standard DNNs (dotted blue line), 
  MscaleDNNs with $k=38\pi$ (dashed orange line), and 
  MscaleDNNs with $k=40\pi$ (solid green line), respectively.}
	\label{re1}
\end{figure}

To examine the accuracy and robustness of different down-scaling mappings, let us 
consider a simple high frequency function $f(x)= \sin(40\pi x)$ with $x \in [0,1]$.
We evaluate three types of neural network classes: standard DNNs, MscaleDNNs, and DNNs with Fourier features. The loss function
is defined by (\ref{eq21_1}). For all neural networks, the sigmoid activation function is 
employed. The learning rate starts at 0.01 and 
decays by a factor of 0.9 every 1,000 steps 
for 100,000 epochs. 
In the case of MscaleDNNs, the down-scaling mapping is defined as 
\begin{equation}
  \Phi(x) = k x,
  \label{fe1} 
\end{equation}
where $k=38\pi$ or $40\pi$. 
We first conduct two simulations using the standard DNNs 
and MscaleDNNs and report the relative $L_2$ 
errors in Figure \ref{re1}.
Based on these numerical results, we conclude that 
the MscaleDNNs with down-scaling mappings 
can effectively reduce the fitting error for high frequency 
functions, which is consistent with the analysis 
error estimate provided in Theorem \ref{thm1}. 
Moreover, the relative $L_2$ error can 
be further minimized if the exact 
high frequency $40\pi$ of $f(x)$ is used in 
constructing the down-scaling mapping.\par 

\begin{figure}
	\begin{minipage}[t]{0.485\textwidth}
		\centering
		\includegraphics[scale=0.31]{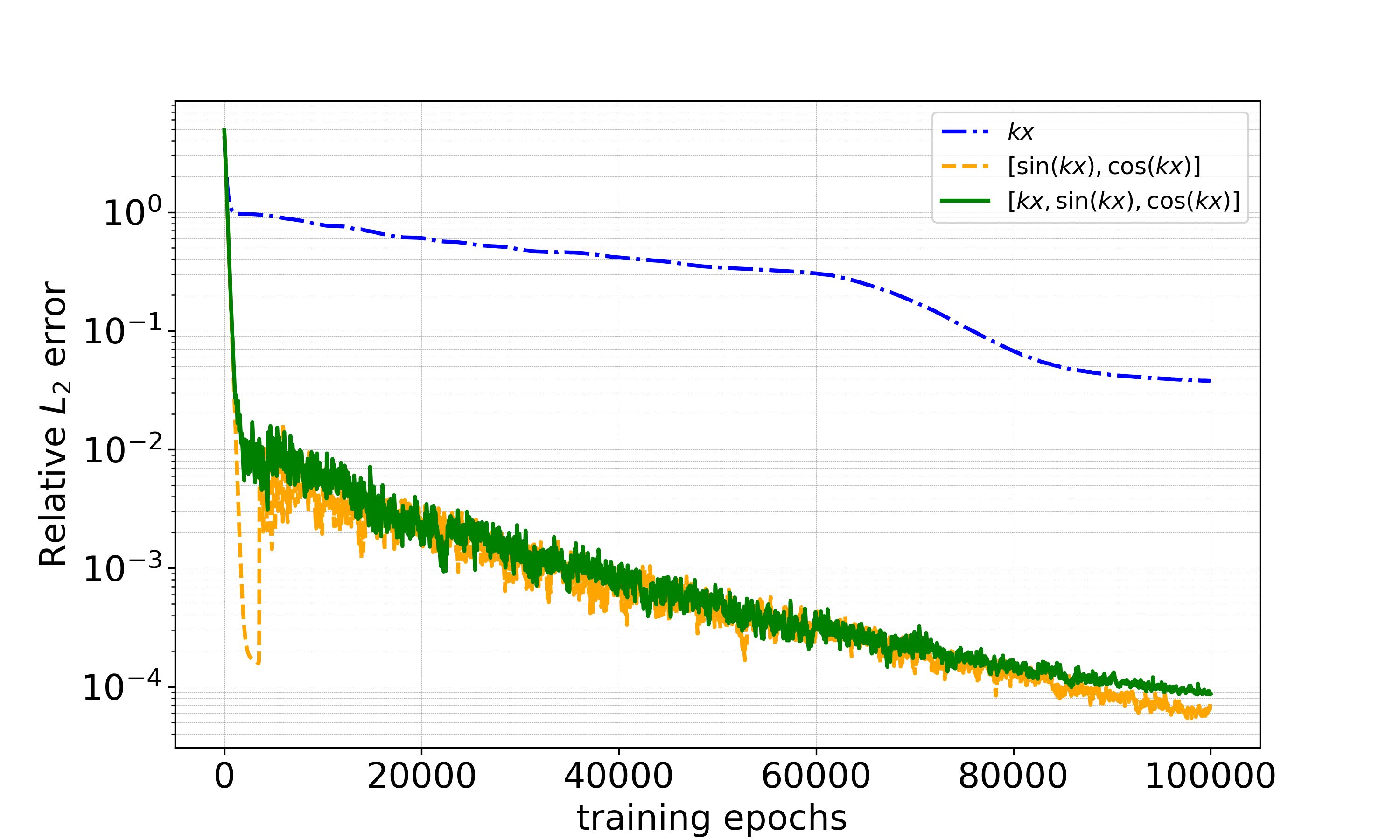}
    \subcaption{$k=40\pi$}
	\end{minipage}%
	\begin{minipage}[t]{0.485\textwidth}
		\centering
		\includegraphics[scale=0.31]{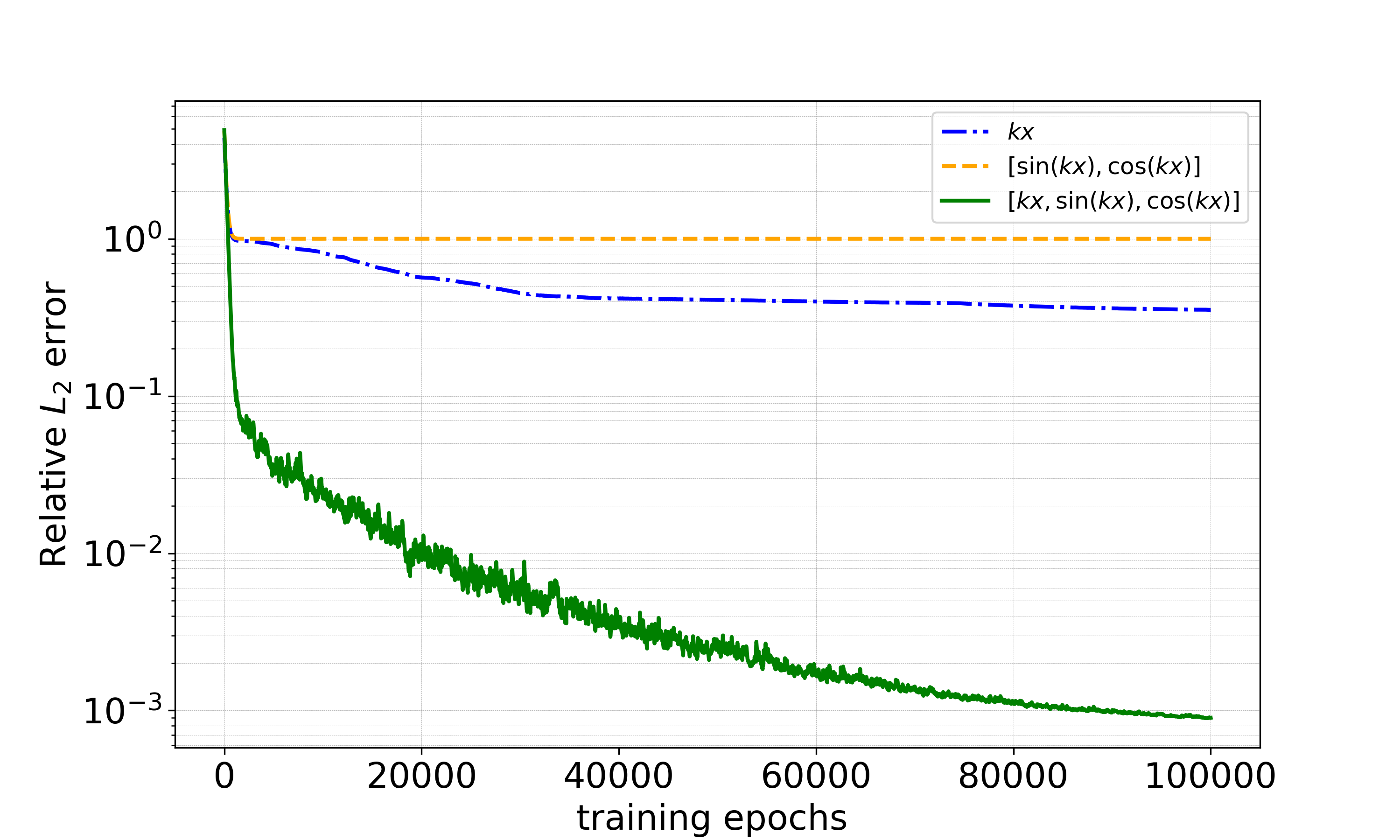}
    \subcaption{$k=38\pi$}
	\end{minipage}
  \caption{(a) $k=40\pi$, (b) $k=38\pi$. The relative $L_2$ error at each training epoch: MscaleDNNs (dotted blue line), 
  DNNs with Fourier feature (dashed orange line), and 
  DNNs with our proposed hybrid feature embedding (solid green line) for different $k$, respectively.}
  \label{re2}
\end{figure}

Next, let us consider the Fourier feature mapping with the following form:
\begin{equation}
  \Phi[k](x) = \begin{bmatrix}
    \sin(k x)\\
    \cos(k x)
  \end{bmatrix},
\end{equation}
where $k=38\pi$ and $40\pi$, 
respectively. The relative $L_2$ errors between $f(x)$ 
and the neural network function $f_{\text{net}}(x)$ 
generating by MscaleDNNs and DNNs with 
Fourier features are shown in Figure \ref{re2}. When $k$ matches the frequency of $f(x)$ (i.e., $k=40\pi$),  DNNs with Fourier 
features can achieve a highly accurate approximation of $f(x)$. However, when $k$ is equal to $38\pi$, the 
performance of DNNs with Fourier features is even worse 
than that of standard DNNs.\par 

\begin{figure}[htbp]
	\centering
	\includegraphics[width=0.8\linewidth]{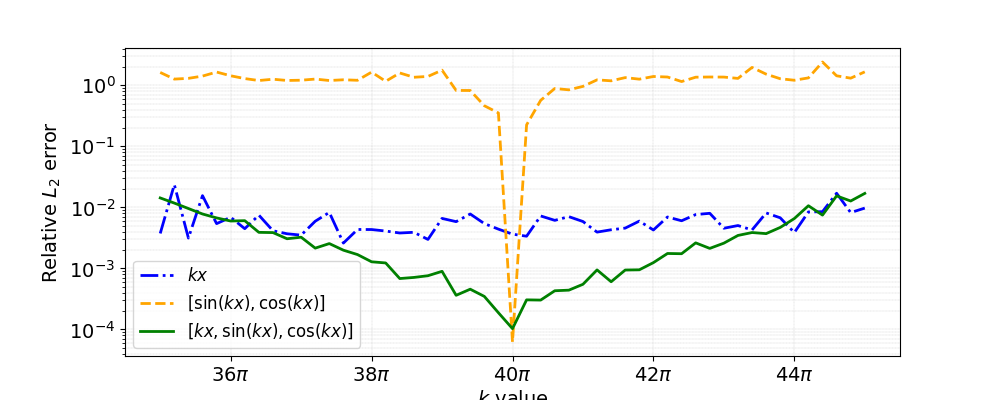}
	\caption{The relative $L_2$ errors after 100,000 training epochs: MscaleDNNs (dotted blue line), 
  DNNs with Fourier embedding (dashed orange line), and 
  DNNs with newly proposed hybrid feature embedding (solid green line) 
  for $k\in [35\pi, 45\pi]$, respectively.}
	\label{re3}
\end{figure}

To understand the performance of DNNs with different Fourier features, 
let us define an auxiliary function
\begin{equation*}
  F(y) = \sin(\hat{k}\arcsin(y)),
\end{equation*}
where $\hat{k}=\frac{40\pi}{k}$
and $y=\sin(k x)=\sin(\frac{40\pi}{\hat{k}}x)\in [-1,1]$.
When $\hat{k}=1$, 
it is clear that 
\begin{equation*}
  f(x)= F\left(\sin(40\pi x)\right):=F(y)=y=\sin(40\pi x),
\end{equation*}
which satisfies the assumptions given in Theorem \ref{thm_co1}. This explains why 
DNNs with Fourier features $\Phi[40\pi](x)$ can achieve an accurate approximation of $f(x)$. 
For $|\hat k|\neq 1$, we have 
\begin{equation*}
  F'(y) = \hat{k} \cos(\hat{k}\arcsin(y))\frac{1}{\sqrt{1-y^2}}.
\end{equation*}
Due to the following limit
\begin{equation*}
  \lim\limits_{y\to 1} \left|\hat{k} \cos( \hat{k} \arcsin(y))\frac{1}{\sqrt{1-y^2}}\right| = \infty,
\end{equation*}
we have $\|F\|_{C^{1}[-1,1]}=\infty$. Consequently, the assumption 
in Theorem \ref{thm_co1} does not hold, and the fitting error estimate becomes invalid even if $k$ is close to 1.
To verify it, we run several simulations with different $k$ values in the range $[35\pi, \ 45 \pi]$, and 
the relative $L_2$ errors are reported in Figure \ref{re3}. 
According to these numerical results, we observe that DNNs with Fourier 
features is more accurate than MscaleDNNs when the exact frequency of 
$f(x)$ is used. However, for $k \in [35\pi, 40\pi]$ where $k \neq 40\pi$, 
MscaleDNNs can obtain an 
acceptable accurate approximation of $f(x)$, while the performance of DNNs with Fourier 
features is poor. Therefore, 
MscaleDNNs exhibit greater robustness compared to DNNs 
with Fourier embedding.

To leverage the strengths of both MscaleDNNs and DNNs with Fourier features, we introduce a hybrid feature embedding in this paper. 
For $\bm{x}\in \mathbb{R}^d$, this embedding is defined as: 
\begin{equation}
  \Phi[\bm{k}] (\bm{x}) = \begin{bmatrix}
    \bm{k} \cdot \bm{x} \\
    \cos(\bm{k} \cdot \bm{x})\\
    \sin(\bm{k}\cdot  \bm{x})
  \end{bmatrix}.
  \label{fe3}
\end{equation}
We then use DNNs with this hybrid feature embedding to fit $f(x)$. The 
relative $L_2$ errors at each training epoch for this hybrid feature embedding, with $k=38\pi$ and $40\pi$, are also 
displayed in Figure \ref{re2}. 
For $k=40\pi$, DNNs with the hybrid feature embedding 
can obtain an approximation for $f(x)$ with accuracy comparable to that of DNNs with Fourier feature embedding. For $k=38\pi$, 
the error for DNNs with the hybrid feature embedding is much
less than MscaleDNNs. We also 
apply the DNNs with hybrid feature embedding to fit $f(x)$ for $k\in [35\pi, \ 45\pi]$. 
The numerical results, presented in Figure \ref{re3}, confirm that the newly proposed 
hybrid feature embedding offers advantages in both accuracy and robustness. Consequently, we 
will take the hybrid feature embedding as multiple inputs 
for the frequency-adaptive MscaleDNNs. 

\subsection{Posterior error estimate and frequency capture for high frequency functions}
\label{pf}
We then establish a posterior error estimate to 
capture the frequency information of the fitted function $f$. 
Let $f\in C[0,1]^d$ and $f_{\text{net}}\in C[0,1]^d$ 
denote the output of a neural network after several training epochs. The Fourier series of 
$f_{\text{net}}$ is given by 
\begin{equation*}
  f_{\text{net}}(\bm{x}) = \sum_{\bm{k}\in \mathbb{Z}^d} \hat{f}_{\text{net}, \bm{k}} \mathrm{e}^{\mathrm{i}2\pi \bm{k}\cdot \bm{x}},
\end{equation*}
where $\mathbb{Z}^d=\{(k_1,\ldots,\ k_d)\mid \ k_1,\ldots,\ k_d \in \mathbb{Z}\}$ and
\begin{equation*}
  \hat{f}_{\text{net}, \bm{k}} = \int_{[0,1]^d} f_{\text{net}}(\bm{x})\cdot \mathrm{e}^{-\mathrm{i}2\pi \bm{k}\cdot \bm{x}} \ \mathrm{d}\bm{x}.
\end{equation*}
Furthermore, if $f_{\text{net}}$ is a band-limited function, then 
\begin{equation*}
  f_{\text{net}}(\bm{x}) = \sum_{\bm{k}\in \mathbb{B}} \hat{f}_{\text{net}, \bm{k}} \mathrm{e}^{\mathrm{i}2\pi \bm{k}\cdot \bm{x}},
\end{equation*}
where $\hat{f}_{\text{net},\bm{k}}$ can be calculated through the Discrete 
Fourier Transform (DFT) and $\mathbb{B}$ represents the compact support.
Let us select the $N_0$ Fourier coefficients with the largest modulus among them as $\hat{f}_{\text{net}, \bm{k}_1},\dots,\hat{f}_{\text{net},\bm{k}_{N_0}}$. The value 
of $N_0$ is large enough such that 
\begin{equation}
  \label{get_n}
  \sum_{j=1}^{N_0}|\hat{f}_{\text{net},\bm{k}_j}|^2 \geqslant  (1-\delta)\|f_{\text{net}}\|^2_{L^2([0,1]^d)} ,
\end{equation}
where $0 \leq \delta < 1$ is a preselected parameter.

Let us assume that
\begin{equation}
  \label{get_err}
  \|f_{\text{net}} - f\|_{L^2([0,1]^d)} \leq \epsilon,
\end{equation}
where $0<\epsilon \ll \|f\|_{L^2([0,1]^d)}$. As reported in 
\cite{liu2020multi}, the assumption (\ref{get_err}) is generally valid for $f_{\text{net}}$ when it is the output of MscaleDNNs, even for fitting a multi-scale function $f$. Based on the assumption 
(\ref{get_err}), we present the posterior error estimate for the Fourier coefficients of $f$.
Let us denote the Fourier series for a band-limited function $f$ as 
\begin{equation*}
  f(\bm{x}) = \sum_{\bm{k}\in \mathbb{B}} \hat{f}_{\bm{k}} \mathrm{e}^{\mathrm{i}2\pi \bm{k}\cdot \bm{x}},
\end{equation*}
where the Fourier coefficients are given by 
\begin{equation*}
  \hat{f}_{\bm{k}} = \int_{[0,1]^d} f(\bm{x})\mathrm{e}^{-\mathrm{i}2\pi \bm{k}\cdot \bm{x}} \ \mathrm{d}\bm{x}.
\end{equation*}
By setting $\hat{\mathbb{B}} = \{\bm{k}_1, \dots, \bm{k}_{N_0}\}$, we have 
\begin{equation}
\label{new:eq2}
  \begin{split}
    \sum_{j=1}^{N_0} |\hat{f}_{\text{net},\bm{k}_j} - \hat{f}_{\bm{k}_j}|^2 & \leq \sum_{\bm{k}\in \mathbb{B}} |\hat{f}_{\text{net},\bm{k}} - \hat{f}_{\bm{k}}|^2 \\
    &  = \| f_{\text{net}} - f\|^2_{L^2([0,1]^d)}\leq \epsilon^2, 
  \end{split}
\end{equation}
which implies that $|\hat{f}_{\text{net},\bm{k}_j} - \hat{f}_{\bm{k}_j}|\leq \epsilon$ holds for 
all $\bm{k}_j\in \hat{\mathbb{B}}$. Therefore, the frequency coefficients $\hat{f}_{\text{net},\bm{k}_j}$ with $\bm{k}_j\in \hat{\mathbb{B}}$
provide a good approximation for $\hat{f}_{\bm{k}_j}$. Furthermore, we have 
\begin{equation}
\label{neweq:1}
  \begin{split}
    \sum_{j=1}^{N_0} |\hat{f}_{\bm{k}_j}|^2 & = \sum_{j=1}^{N_0} \left|\hat{f}_{\bm{k}_j}- \hat{f}_{\text{net},\bm{k}_j} +\hat{f}_{\text{net},\bm{k}_j}\right|^2 \\
    & 
    \geq \left[\left(\sum_{j=1}^{N_0} |\hat{f}_{\text{net},\bm{k}_j}|^2\right)^{\frac{1}{2}} - \left(\sum_{j=1}^{N_0} \left|\hat{f}_{\bm{k}_j}- \hat{f}_{\text{net},\bm{k}_j}\right|^2\right)^{\frac{1}{2}} \right]^2 \\ 
    & \geq \big[(1-\delta)^{\frac{1}{2}}\|f_{\text{net}}\|_{L^2([0,1]^d)} - \epsilon\big]^2 \\
    & \geq \big[(1-\delta)^{\frac{1}{2}}\|f\|_{L^2([0,1]^d)} - [(1-\delta)^{\frac{1}{2}}+1]\epsilon\big]^2,
  \end{split}
\end{equation}
which implies that $\hat{f}_{\bm{k}_j}$ with $\bm{k}_j\in \hat{\mathbb{B}}$ dominate the contribution to 
$L_2$ norm of function $f$. In \eqref{neweq:1}, the first inequality follows from the triangle inequality of norm, and the last two inequalities are justified by $0<\epsilon \ll \|f\|_{L^2([0,1]^d)}$, \eqref{get_err}, and \eqref{new:eq2}.
For $\bm{k}_j \in \hat{\mathbb{B}}$, by constructing a new neural network with 
multiple inputs $\Phi[\bm{k}_j](\bm{x})$ defined in (\ref{fe3}), we can obtain 
a new neural network function $f_{\text{net}}^{\text{new}}$. According to Theorem \ref{thm2}, 
the distance between $f_{\text{net}}^{\text{new}}$ and $f$ is independent on the high frequency components 
of $f$, which is typically less than $\|f_{\text{net}} - f\|_{L^2([0,1]^d)}$.  \par 

In summary, we propose an approach to reduce the fitting error for high 
frequency function $f$ through posterior frequency capture. Initially, we 
pre-train a neural network using either a single input or multiple random
feature inputs to fit high frequency function $f$, ensuring that (\ref{get_err}) holds. 
Next, we apply the DFT to identify the $N_0$ Fourier coefficients with the largest modulus from the pre-trained neural network function $f_{\text{net}}$. These dominant frequencies are then used to construct a new neural network with multiple inputs. Finally, we train this new network to achieve a more accurate function approximation. This process can be iteratively repeated until convergence. Based on these observations, we introduce a frequency-adaptive MscaleDNNs and use it to solve PDEs with multi-scale solutions, in the next subsection.

\subsection{Frequency-adaptive MscaleDNNs}
\label{fr}
For a bounded domain $\Omega \in \mathbb{R}^d$, let us consider 
the following PDE:
\begin{equation}
	\label{eq7}
	\left\{ 
		\begin{aligned}
			&\mathcal{N}(\bm{x};u(\bm{x})) = 0, \quad \bm{x}\in \Omega, \\
			&\mathcal{B}(\bm{x};u(\bm{x})) = 0, \quad \bm{x}\in \partial \Omega,
		\end{aligned}
	\right.
\end{equation}
where $\mathcal{N}$ is the differential operator, 
$\mathcal{B}$ is the boundary operator, 
and $u(\bm{x})$ is the multi-scale unknown solution.
For convenience of description, let us 
assume $\Omega = (0,1)^d$. Inspired by PINNs, we seek an approximate 
solution $u_{\text{net}}(\bm{x};\theta)$ for (\ref{eq7})
by solving the following soft constrained 
optimization problem:
\begin{equation}
	\min_{\theta \in \Theta} \mathcal{L}(\theta)=\mathop{\text{min}}\limits_{\theta \in \Theta} \  w_r \cdot \mathcal{L}_r(\theta) + w_b \cdot \mathcal{L}_b(\theta),
  \label{loss1}
\end{equation}
where 
\begin{equation*}
	\mathcal{L}_r(\theta) = \sum_{i=1}^{N_r}\big{|}\mathcal{N}(\bm{x}_r^i;u_{\text{net}}(\bm{x}_r^i;\theta))\big{|}^2 \quad \text{and} \quad \mathcal{L}_b(\theta) = \sum_{i=1}^{N_{b}}\big{|}\mathcal{B}(\bm{x}_{b}^i;u_{\text{net}}(\bm{x}_{b}^i;\theta))\big{|}^2.
\end{equation*}
Here $\Theta$ is the parameter space, and $\{w_r, \ w_b\}$ are weights used
to balance the domain loss $\mathcal{L}_r(\theta)$ and the boundary 
loss $\mathcal{L}_b(\theta)$. 
In (\ref{loss1}), $\{\bm{x}_r^i\}_{i=1}^{N_r}$ and $\{\bm{x}_{b}^i\}_{i=1}^{N_{b}}$ 
represent the sample points within the domain $\Omega$ and  on the boundary $\partial \Omega$, 
respectively. 
The optimal parameters $\theta$ are usually 
found by minimizing the loss function (\ref{loss1}) through 
a stochastic gradient-based algorithm.\par 

The parameter space $\Theta$ depends on the structure of neural networks. In order to 
obtain a multi-scale solution for (\ref{eq7}), we first employ 
the MscaleDNNs to approximate the unknown 
solution $u(\bm{x})$. In this setup, the 
down-scaling parameters $a_i $ belong to $ \mathbb{B}_0= \{2^0, \, \ldots, \, 2^{M_0 -1}\}$,  
where $M_0$ denotes the number of sub networks. 
After $T_0$ training steps, a preliminary 
solution $u_{\text{net},0}(\bm{x};\theta_0^*)$ is obtained.
Due to the down-scaling mapping 
in the first hidden layer of the MscaleDNNs, 
the relative $L_2$ error $\|u_{\text{net},0}(\bm{x};\theta_0^*) - u(\bm{x})\|_{L^2(\Omega)}/\|u(\bm{x})\|_{L^2(\Omega)}$
is typically smaller compared to that of a standard neural network.
As reported in \cite{liu2020multi}, the error is usually on the order 
of $10^{-2}\thicksim 10^{-1}$. Therefore, the inequality (\ref{get_err})
will hold for a small $\epsilon$.

With the values of $u_{\text{net},0}(\bm{x};\theta_0^*)$ on the $N^d$
uniform mesh points in $\Omega$, we can perform DFT for $u_{\text{net},0}(\bm{x};\theta_0^*)$ to 
obtain the Fourier coefficients $\hat{u}_{\text{net},0,\bm{k}}$, 
where $\bm{k}\in \mathbb{B}$ and $|\mathbb{B}|\leq N^d$.
Let us denote $\zeta = \max_{\bm{k} \in \mathbb{B}} \big|\hat{u}_{\text{net},0,\bm{k}}\big|$.
We then introduce a parameter $\lambda \in (0,1)$ to select all frequencies such that 
\begin{equation}
  \label{pick_k}
  |\hat{u}_{\text{net},0,\bm{k}}| > \lambda \zeta.
\end{equation}
The selected frequencies are denoted as $\bm{k}_j$ with 
$j = 1, \ldots, N_0$, ordered such that $|\bm{k}_j|_1 < |\bm{k}_l|_1$ for $j<l$. For the $0\leq \delta < 1$, by choosing 
parameter $\lambda$ appropriately, we can ensure that 
\begin{equation}
  \sum_{i=1}^{N_0}|\hat{u}_{\text{net},0,\bm{k}_i}|^2 \geqslant  (1-\delta)\|u_{\text{net},0}(\bm{x};\theta_0^*)\|^2_{L^2([0,1]^d)}.
  \label{get_delta}
\end{equation}
 The choice of $\lambda$ should depend on the distribution 
of $\hat{u}_{\text{net},0,\bm{k}}$. For $u_{\text{net},0}$ with sparsely distributed Fourier coefficients, 
$\lambda$ can be set to a small value, such as $\lambda=0.01$. Otherwise, $\lambda$ should be 
set to a relatively large value.
With the selected frequency set $\mathbb{B}_1 =\{\bm{k}_1, \ldots, \ \bm{k}_{N_0}\}$, we 
adaptively adjust the neural 
network according to the following two criteria.

\begin{itemize}
  \item {\bf Criterion A}: If $N_0 \leqslant M_0$, we construct a new neural network 
  with $N_0$ sub networks. The input of the $j$-th sub network is set as 
  $\Phi[\bm{k}_j](\bm{x})$, as defined in (\ref{fe3}). The output 
  of the newly constructed neural network $y$ is then given by 
  \begin{equation}
    y = \sum_{j=1}^{N_0} h_j y_j,
    \label{out1}
  \end{equation}
  where $h_j=\hat{u}_{\text{net},0,\bm{k}_j}$ and $y_j$ is the output of the $j$-th sub network. Here, the 
  values of $\hat{u}_{\text{net},0,\bm{k}}$ 
  are used to accelerate the convergence of the newly constructed neural network.

  \item {\bf Criterion B}: For $N_0 > M_0$, 
  rather than increasing the number of
  sub networks, we suggest incorporating  multiple 
  frequency features to one sub network.
  Let us denote $\mathbb{B}_1=\{\bm{k}_1,\  \bm{k}_2 \ldots, \  \bm{k}_{N_0}\}$.
  We divide $\mathbb{B}_1$ into $M_0$ subsets $\mathbb{B}_1^j$ with 
  $j=1,\ldots,\ M_0$. Typically, $\mathbb{B}_1^1$ is set as 
  $\mathbb{B}^1_1 = \{\bm{k}_1, \ldots, \ \bm{k}_{\lfloor \frac{N_0}{M_0} \rfloor}\}$, ..., 
  and $\mathbb{B}_1^{M_0}$ is set as 
  $\mathbb{B}_1^{M_0}=\{\bm{k}_{(M_0-1)\lfloor \frac{N_0}{M_0} \rfloor + 1}, \ldots, \ \bm{k}_{N_0}\}$, respectively.
  The inputs for the $j$-th sub network are $\Phi[\bm{k}](\bm{x})$ with $\bm{k}\in \mathbb{B}^j_1$. 
  The output of the newly constructed neural network is defined as 
  \begin{equation}
    y = \sum_{j=1}^{M_0} h_j y_j,
    \label{out2}
  \end{equation}
  where $y_j$ is the output of the $j$-th sub network. In this work, we take 
  $h_j = \sum_{\bm{k} \in \mathbb{B}^j_1} \big|\hat{u}_{\text{net}, 0,\bm{k}}\big|$ 
  to accelerate the convergence of the newly constructed neural network.
\end{itemize}

By iteratively applying the posterior frequency capture and neural network adjustment processes, we develop the frequency-adaptive MscaleDNNs algorithm, which is summarized in Algorithm \ref{al}.  We propose two criteria for terminating the adaptive iteration.  The first criterion is to fix the maximal number of adaptations $I$. The second criterion involves comparing the captured frequency features sets $\mathbb{B}_{It}$ and $\mathbb{B}_{It+1}$. 
If $\mathbb{B}_{It} = \mathbb{B}_{It+1}$, the iteration process is halted.

\begin{remark}
  The output of the newly constructed neural networks belongs to the feature mapping
  network class $\widetilde{\mathcal{F}}(N, L,\iota, \delta )$. We denote the output as:
  \begin{equation*}
    u_{\text{net}} = WG + b,
  \end{equation*}
  where $G = (y_1, \ y_2, \ldots, \ y_{N_0})^T$ or $G = (y_1, \ y_2, \ldots, \ y_{M_0})^T$. 
  In this work, we set $W=(h_1, \ h_2, \ldots)$ and $b=0$ to reduce the learning complexity. 
  Compared to the learnable parameters $W$ and $b$, numerical 
  simulations reported in section \ref{po} show that
  fixing $W$ and $b$ provides advantages 
  in both accuracy and efficiency. 
\end{remark}

\begin{algorithm}[htbp]
	\caption{Frequency-adaptive MscaleDNNs algorithm} 
  \label{al}
	\hspace*{0.02in} {\bf Require:}
	The number of sub networks $M_0$, initial feature set $\mathbb{B}_0 = \{2^0, \dots, 2^{M_0-1}\}$,
	total adaptive steps $I$ and parameter $\lambda$.
	\begin{algorithmic}[1]
    \STATE $It \leftarrow 0$
    \WHILE{$It \leqslant  I$}
		\STATE Train the MscaleDNNs with feature set $\mathbb{B}_{It}$ to obtain an approximate solution 
    $u_{\text{net},It}$ for the given PDE. 

		\STATE Perform DFT on $u_{\text{net},It}$ to derive the 
    Fourier coefficients $\hat{u}_{\text{net},It,\bm{k}}$.

    \STATE For all $\bm{k} \in \mathbb{B} $, if $|\hat{u}_{\text{net},It,\bm{k}}| > \lambda \max_{\bm{k} \in \mathbb{B}} |\hat{u}_{\text{net},It,\bm{k}}|$, include $\bm{k}$ in $\mathbb{B}_{It+1}$.
    \IF{$\mathbb{B}_{It} = \mathbb{B}_{It+1}$}
    \STATE Break
    \ENDIF
    \STATE Adjust MscaleDNNs following the criteria A or B.
		\ENDWHILE
	\end{algorithmic}
  \hspace*{0.02in} {\bf Output:}
  output $u_{\text{net},I}$ or $u_{\text{net},It}$ when $\mathbb{B}_{It} = \mathbb{B}_{It+1}$.
\end{algorithm}

\section{Numerical examples}
\label{nu}
In this section, we evaluate the performance 
of the proposed frequency-adaptive 
strategy in solving multi-scale PDEs. 
The network initialization 
follows the Glorot normal scheme 
\cite{glorot2010understanding}. All networks are 
trained using
the Adam optimizer \cite{kingma2014adam} with defaulting settings. We employ an initial learning rate of $0.01$ or $0.001$.
In all simulations, the activation function used for MscaleDNNs is 
the Softened Fourier Mapping (SFM) activation function, 
defined as $0.5\sin(x) + 0.5\cos(x)$. This activation function 
produces similar results to the sigmoid function but converges more 
quickly \cite{li2023subspace}.
The relative $L_2$ error are defined as 
\begin{equation*}
  \text{rel error} = \frac{\|u-u_{\text{net},It}\|_{L^2(\Omega)}}{\|u\|_{L^2(\Omega)}},
\end{equation*}
where $u$ is the reference solution and $u_{\text{net}, It}$ is the output of 
frequency-adaptive MscaleDNNs.\par

\subsection{Poisson equation}
\label{po}
\begin{figure}[H]
  \centering
  \includegraphics[width=1.0\linewidth]{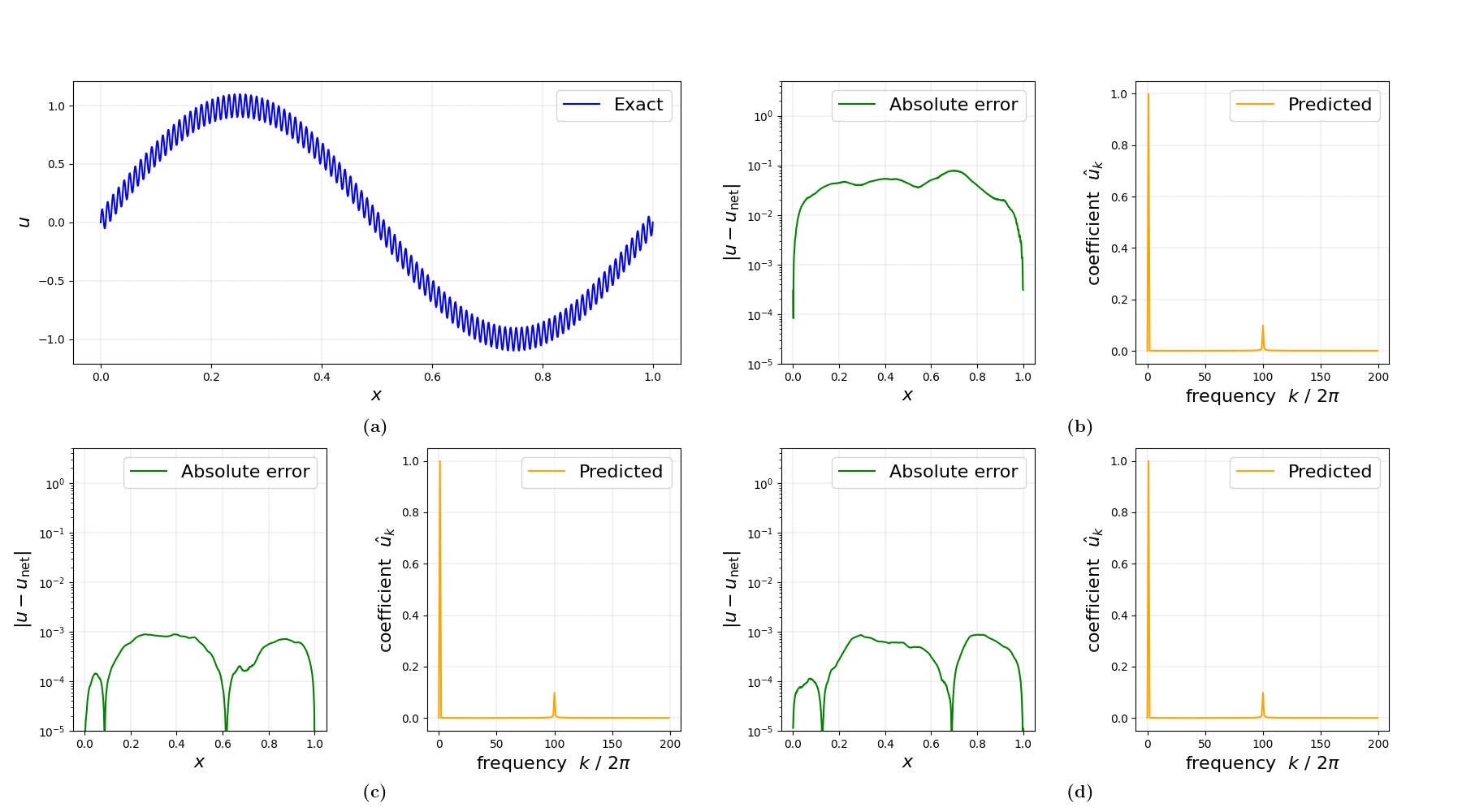}
  \caption{Poisson equation. (a) The exact solution $u(x)$.
  (b) Step $It=0$. Left: absolute error $|u - u_{\text{net},0}|$, right: distribution of Fourier coefficients.
  (c) Step $It=1$. Left: absolute error $|u - u_{\text{net},1}|$, right: distribution of Fourier coefficients. 
  (d) Step $It=2$. Left: absolute error $|u - u_{\text{net},2}|$, right: distribution of Fourier coefficients.}
  \label{po_fig}
\end{figure}

We consider the following one-dimensional Poisson equation:
\begin{equation}
	-\bigtriangleup u(x) = f(x), \quad u(0)=u(1)=0.
  \label{po_eq}
\end{equation}
By setting $f(x)=4\pi^2\sin(2\pi x) + 4,000\pi^2\sin(200\pi x)$, the 
multi-scale PDE (\ref{po_eq}) exists an exact solution 
$u(x)=\sin( 2\pi x) + 0.1 \sin(200 \pi x)$. 
We employ the frequency-adaptive MscaleDNNs with $I=4$ to numerically solve 
the Poisson equation \eqref{po_eq}. Initially, the MscaleDNNs include 
$6$ sub networks with multiple frequency inputs $\mathbb{B}_0 = \{2^0, \ldots, \ 2^5\}$, and the 
parameters for each sub network are $[1,100,100,100,1]$. 
The learning rate 
decays exponentially at a rate of 0.9 every 500 
training iterations.
At the $It$-th adaptive step, we train the MscaleDNNs 
100,000 training epochs to obtain the neural network solution $u_{\text{net},It}$. 
The absolute error $|u - u_{\text{net},0}|$ is displayed in Figure \ref{po_fig}-(b),
which shows that $u_{\text{net},0}$ is not very accurate. We then use DFT 
to obtain the distribution of Fourier coefficients for $u_{\text{net}, 0}$.
The result, also reported in Figure \ref{po_fig}-(b), is almost consistent with 
the distribution of Fourier coefficients for $u(x)$. By setting $\lambda=0.01$, we obtain the frequency 
feature set $\mathbb{B}_1=\{2\pi, 4\pi, 200\pi, 202\pi\}$.\par 

Using the frequency feature set $\mathbb{B}_1$, we construct 
a new MscaleDNNs following Criterion A. The newly constructed neural network
includes four sub networks, with inputs  
$\Phi[2\pi](x)$, $\Phi[4\pi](x)$, $\Phi[200\pi](x)$, and $\Phi[202\pi](x)$ for each sub network, respectively.
We then train it to get $u_{\text{net},1}$.
The associated absolute error and distribution of Fourier coefficients are shown 
in Figure \ref{po_fig}-(c). 
With the help of posterior frequency $\mathbb{B}_1$, the absolute error $|u-u_{\text{net},1}|$ is 
nearly tow orders of magnitude smaller than $|u-u_{\text{net},0}|$, demonstrating 
that the frequency-adaptive MscaleDNNs significantly improve the accuracy 
of the standard MscaleDNNs after one adaptive iteration. By comparing the Fourier coefficient 
distributions of $u_{\text{net},0}$ and $u_{\text{net},1}$, we find 
that the distribution of Fourier coefficients for $u_{\text{net},1}$ is more consistent with 
that of the exact solution $u(x)$. 
Applying DFT to $u_{\text{net},1}$, 
we obtain the frequency feature 
set $\mathbb{B}_2=\{2\pi, 200\pi, 202\pi\}$ for the second adaptive step. 
Training the neural network again yields a new output $u_{\text{net},2}$. 
The absolute error $|u-u_{\text{net},2}|$ and the distribution of Fourier coefficients for 
$u_{\text{net},2}$ are displayed in Figure \ref{po_fig}-(d). By performing 
DFT on $u_{\text{net},2}$, we get the frequency set $\mathbb{B}_3=\{2\pi, 200\pi, 202\pi\}$.
The simulation is stopped at $It=2$ as the stopping 
condition $\mathbb{B}_2=\mathbb{B}_3$ is satisfied.
Since we only use 1,000 mesh points in the DFT, there exists a  discrepancy between the captured frequencies and the exact solution's frequencies due to the DFT error. This discrepancy can be eliminated by increasing the number of mesh points in the DFT. 
The relative 
$L_2$ error of each adaptive iteration is listed 
in Table \ref{table1}. 
The predicted solution of the frequency-adaptive MscaleDNNs 
demonstrates excellent agreement with the exact solution, 
yielding a relative $L_2$ error of 7.196e-04.  

In equation (\ref{out1}) and (\ref{out2}), the parameters $h_j$ are 
set based on the frequency information of the output at the previous 
adaptive iteration. These parameters can also be made learnable. In Table \ref{table2}, 
we compare the relative errors of frequency-adaptive MscaleDNNs with the 
fixed parameters $h_j$ and learnable parameters $h_j$, finding that the errors are essentially the same. By fixing $h_j$, the total number 
of parameters of the frequency-adaptive MscaleDNNs decreases. 
Therefore, to reduce computational cost during training, we always 
fix the values of $h_j$ in this work.

\begin{table}[H]
  \begin{center}
  \caption{Poisson equation (\ref{po_eq}). Relative $L_2$ error at each adaptive step.}
  \begin{tabular}{cccccc}
    \hline
    Adaptive step ${It}$ & 0 & 1 & 2\\
    \hline 
    Fixed parameters $h_j$ & 6.227e-02 & 7.795e-04& 7.196e-04 \\
    \hline
    Learnable parameters $h_j$ & 6.227e-02 & 8.942e-04& 8.228e-04\\
    \hline
    \label{table2}
  \end{tabular}
\end{center}
\end{table}

\begin{table}
  \begin{center}
  \caption{Relative $L_2$ error at each adaptive step for different equations.}
  \begin{tabular}{cccccc}
    \hline
    Adaptive step ${It}$ & 0 & 1 & 2 & 3 & 4\\
    \hline 
    Poisson equation (\ref{po_eq}) & 6.227e-02 & 7.795e-04& 7.196e-04& \textemdash & \textemdash \\
    \hline
    Heat equation (\ref{heat_eq})  & 2.673e-01 & 5.366e-04& 3.931e-04& \textemdash & \textemdash\\
    \hline
    Wave equation (\ref{wave_eq})  & 4.256e-01 & 6.325e-03& 1.889e-03& \textemdash& \textemdash \\
    \hline
    Schr$\ddot{\text{o}}$dinger equation (\ref{sc_eq})  & 9.075e-01 & 1.280e-02& 7.603e-03& 5.055e-03&  7.778e-03\\
    \hline
    \label{table1}
  \end{tabular}
\end{center}
\end{table}

\subsection{Heat equation}
\label{he}
To test the performance of the newly proposed frequency-adaptive MscaleDNNs for 
evolving multi-scale problems, we  
consider a one-dimensional heat equation:
\begin{equation}
	\begin{split}
		&u_t(x,t) = \frac{1}{(500 \pi)^2} u_{xx}(x,t), \quad (x, t) \in (0,1) \times (0,1],\\
		&u(x, 0) = \sin(500 \pi x), \quad x\in(0, 1),\\
		&u(0, t) = u(1, t)= 0, \quad t\in(0,1].
	\end{split}
  \label{heat_eq}
\end{equation}
The exact solution $u(x, t)$ for the heat equation (\ref{heat_eq}) is:
\begin{equation*}
  u(x,t)=\mathrm{e}^{-t}\sin(500\pi x).
\end{equation*}
We employ the frequency-adaptive MscaleDNNs 
with $I=4$ to numerically solve 
the multi-scale PDE (\ref{heat_eq}). 
Following the idea of \cite{lagaris1998artificial}, we 
enforce the initial value 
condition into our neural network, eliminating the need for a loss function 
associated with the initial condition. 
This not only ensures more accurate 
satisfaction of the initial condition 
but also facilitates the rapid 
identification of frequency features.

At $It=0$, we solve the heat equation using the 
MscaleDNNs, which consist of 6 sub networks with a multiple inputs set $\mathbb{B}_0=\{2^0, \ldots, \ 2^5\}$, where 
the parameters for each sub network are  $[1,100,100,100,1]$. 
The learning rate 
decays exponentially at a rate of 0.9 every 500 
training iterations.  
At the $It$-th adaptive step, the neural network solution 
$u_{\text{net},It}$ is obtained by training the network for 50,000 epochs. 
Since the solution $u(x,t)$ lacks periodicity in time, we extend 
it to ensure the periodicity and guarantee the accuracy of DFT. 
In this work, we perform an even expansion of $u(x,t)$ 
and $u_{\text{net},It}$ in time direction. DFT is then performed on domain $(0,1)\times(-1,1)$. \par 

\begin{figure}[htbp]
  \centering
  \includegraphics[width=1.0\linewidth]{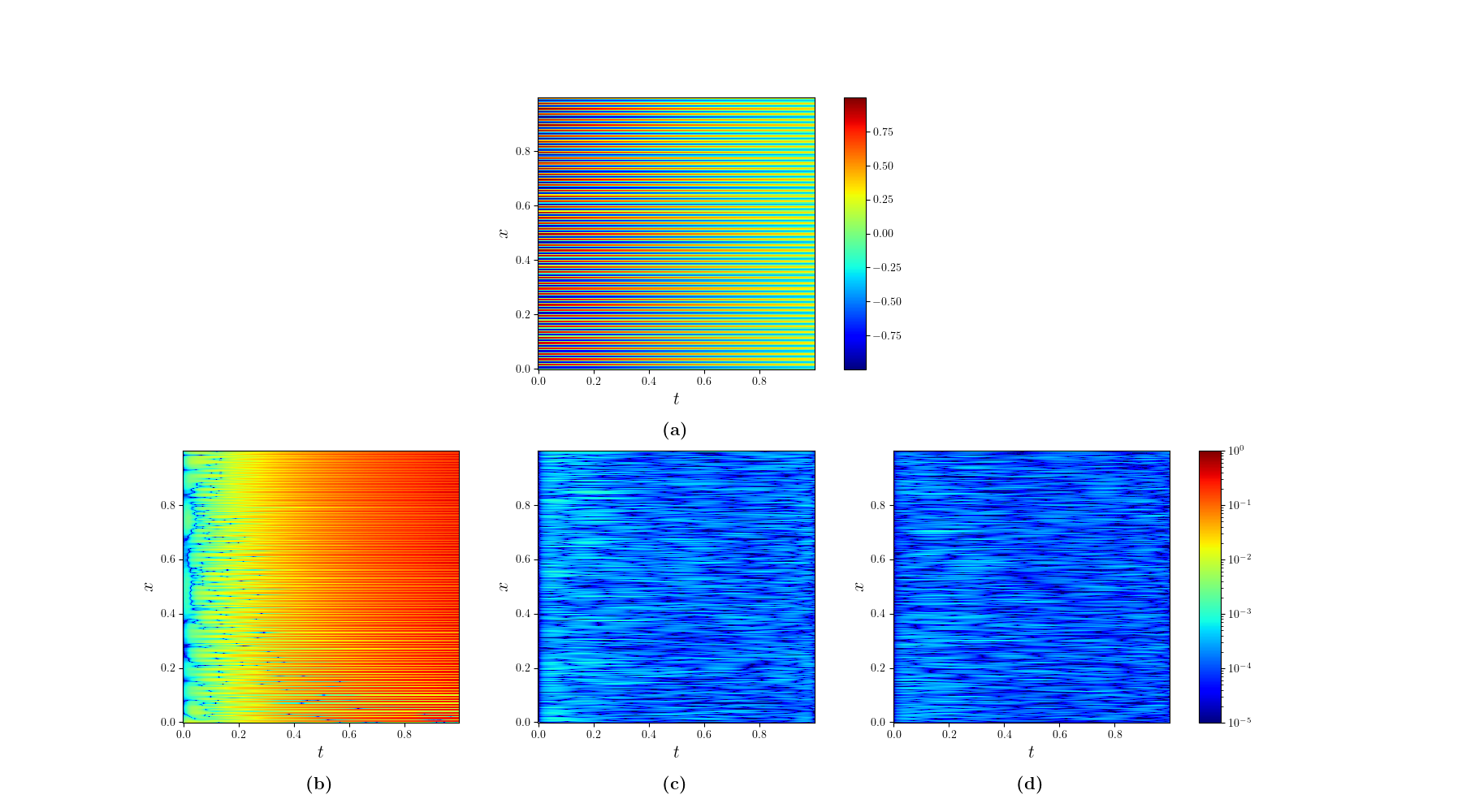}
  \caption{Heat equation. (a) The exact solution $u(x, t)$.
  (b) Step $It=0$, absolute error $|u - u_{\text{net},0}|$.
  (c) Step $It=1$, absolute error $|u - u_{\text{net},1}|$. 
  (d) Step $It=2$, absolute error $|u - u_{\text{net},2}|$.}
  \label{he_err_all}
\end{figure}

\begin{figure}[htbp]
	\centering
  \includegraphics[width=0.8\linewidth]{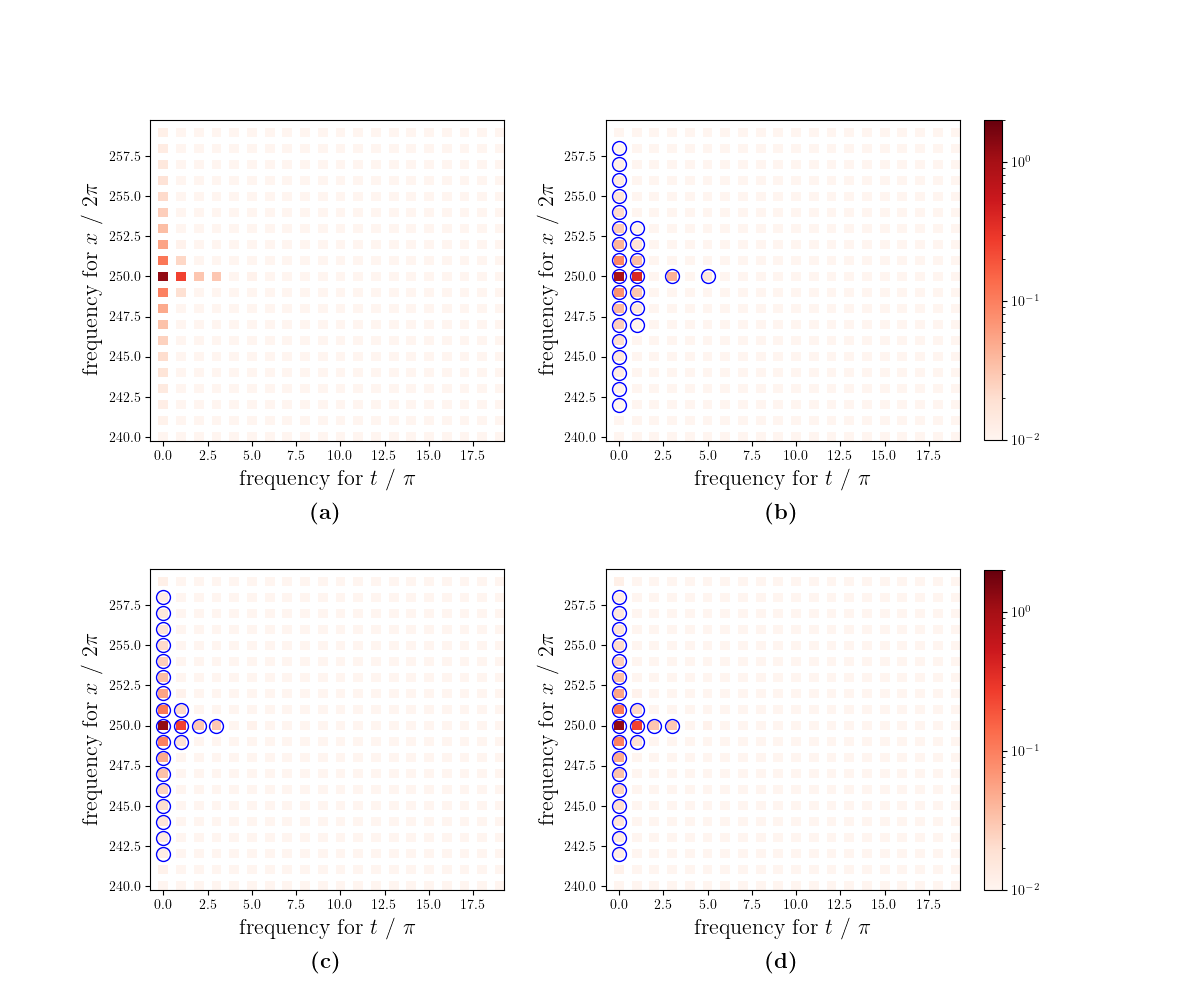}
  \caption{Heat equation. 
  (a) Exact solution's Fourier coefficients, $|\hat{u}_{\bm{k}}|$.
  (b) Step $It=0$, $|\hat{u}_{\text{net},0,\bm{k}}|$.
  (c) Step $It=1$, $|\hat{u}_{\text{net},0,\bm{k}}|$.
  (d) Step $It=2$, $|\hat{u}_{\text{net},0,\bm{k}}|$. In Figures (b)-(d), a blue circle located at $\bm{k}$ means that the frequency $\bm{k}$ has been identified in $\mathbb{B}_{It}$.}
  \label{he_sc}
\end{figure}

The absolute error for MscaleDNNs with the initial multiple inputs set $\mathbb{B}_0$ 
is shown in Figure \ref{he_err_all}-(b), which ranges from approximately $10^{-1}$ to $ 1$. 
By performing the DFT on $u_{\text{net},0}$, we obtain its Fourier coefficients, as reported in Figure \ref{he_sc}-(b). To ensure efficiency, we use $3,000$ mesh points in the $x$-direction to calculate the Fourier coefficients. The Fourier coefficients of exact solution $u(x,t)$, obtained through DFT with $3,000$ mesh points in the $x$-direction, are shown in Figure \ref{he_sc}-(a). Due to accuracy limitations of the DFT with the given number of mesh points, a discrepancy exists between obtained and the exact Fourier coefficients. This discrepancy can be minimized by increasing the number of mesh points used in the DFT. 
By setting $\lambda=0.01$, we identify the frequency feature 
set $\mathbb{B}_1$, {as shown in Figure \ref{he_sc}-(b)}. 
We then construct new MscaleDNNs following Criterion B.
After 50,000 epochs training, we get a new solution $u_{\text{net},1}$, 
with the corresponding absolute error $|u-u_{\text{net},1}|$ displayed in Figure \ref{he_err_all}-(c). 
Since the most important frequencies are included in $\mathbb{B}_1$, the relative error 
$|u-u_{\text{net},1}|$ decreases by several orders of magnitude compared to $|u-u_{\text{net},0}|$. 
According to Figure \ref{he_sc}-(c) and (d), the adaptive iteration continues until it stops at $It=2$, as the stopping condition $\mathbb{B}_2 = \mathbb{B}_3$ is met. 
Table \ref{table1} lists the relative $L_2$ error for each adaptive iteration. The results demonstrate that the predicted solution obtained using the frequency-adaptive MscaleDNNs 
achieves excellent agreement with the
exact solution, yielding a relative $L_2$ error of 3.931e-04.\par

\subsection{Wave equation}
\label{wa}
The third benchmark test case is a one-dimensional wave equation taking the form: 
\begin{equation}
	\begin{split}
		&u_{tt}(x,t) - 25 u_{xx}(x,t)=0, \quad (x, t) \in (0,1) \times (0,1],\\
		&u(0, t) = u(1,t) = 0, \quad t\in (0, 1],\\
		&u(x, 0) = \sin(2\pi x) + \sin(4\pi x), \quad x\in(0, 1),\\
		&u_t(x, 0) = 0, \quad x\in(0,1).
	\end{split}
  \label{wave_eq}
\end{equation}
The exact solution for this equation 
is $u(x,t)=\sin(2\pi x) \cos(10\pi t) + \sin(4\pi x)\cos(20\pi t)$.
As suggested in \cite{daw2022mitigating}, the following loss 
function is applied to solve wave equation (\ref{wave_eq}):
\begin{equation}
  \label{wa_loss}
  \begin{split}
    \mathcal{L}(\bm{\theta}) & = \mathcal{L}_{u}(\bm{\theta}) + \mathcal{L}_{u_t}(\bm{\theta}) + \mathcal{L}_{r}(\bm{\theta}) \\
    & =  \frac{\omega_u}{N_{u}}\sum\limits_{i=1}^{N_{u}}|u_{\bm{\theta}}(x^i_b, t_{u}^i)|^2 + 
     \frac{\omega_{u_t}}{N_{u_t}}\sum\limits_{i=1}^{N_{u_t}}\big|\frac{\partial u_{\bm{\theta}}}{\partial t}(x_{u_t}^i, 0)\big|^2
    +  \frac{\omega_{r}(t_r^i)}{N_r}\sum\limits_{i=1}^{N_r} \big| \frac{\partial^2 u_{\bm{\theta}}}{\partial t^2}(x_r^i, t_r^i) - 
    25 \frac{\partial^2 u_{\bm{\theta}}}{\partial x^2}(x_r^i, t_r^i)\big|^2,
  \end{split}
\end{equation}
where $x^i_b=0$ or $1$, and the weights are set to $\omega_u=1,000$ or $1,000,000$, $\omega_{u_t}=1,000$ or $100,000$. 
The initial value condition is also enforced in our neural network.
In (\ref{wa_loss}), the weight $\omega_{r}(t)$ 
is a continuous gate function defined as 
$\omega_{r}(t)=(1-\tanh(5({t} - \mu )))/2$, where $\mu$
is the scalar shift parameter controlling 
the fraction of time revealed to the model. 
At the $(J+1)$-th training epoch, the shift 
parameter $\mu$ of the causal gate is 
updated to
\begin{equation*}
  \mu_{J+1} =\mu_{J} + 0.002 \mathrm{e}^{-10  \mathcal{L}_r(\theta)},
\end{equation*}
where $\mathcal{L}_r(\theta)$ represents the 
PDE loss at the $J$-th training epoch.\par
\begin{figure}[htbp]
  \centering
  \includegraphics[width=1.0\linewidth]{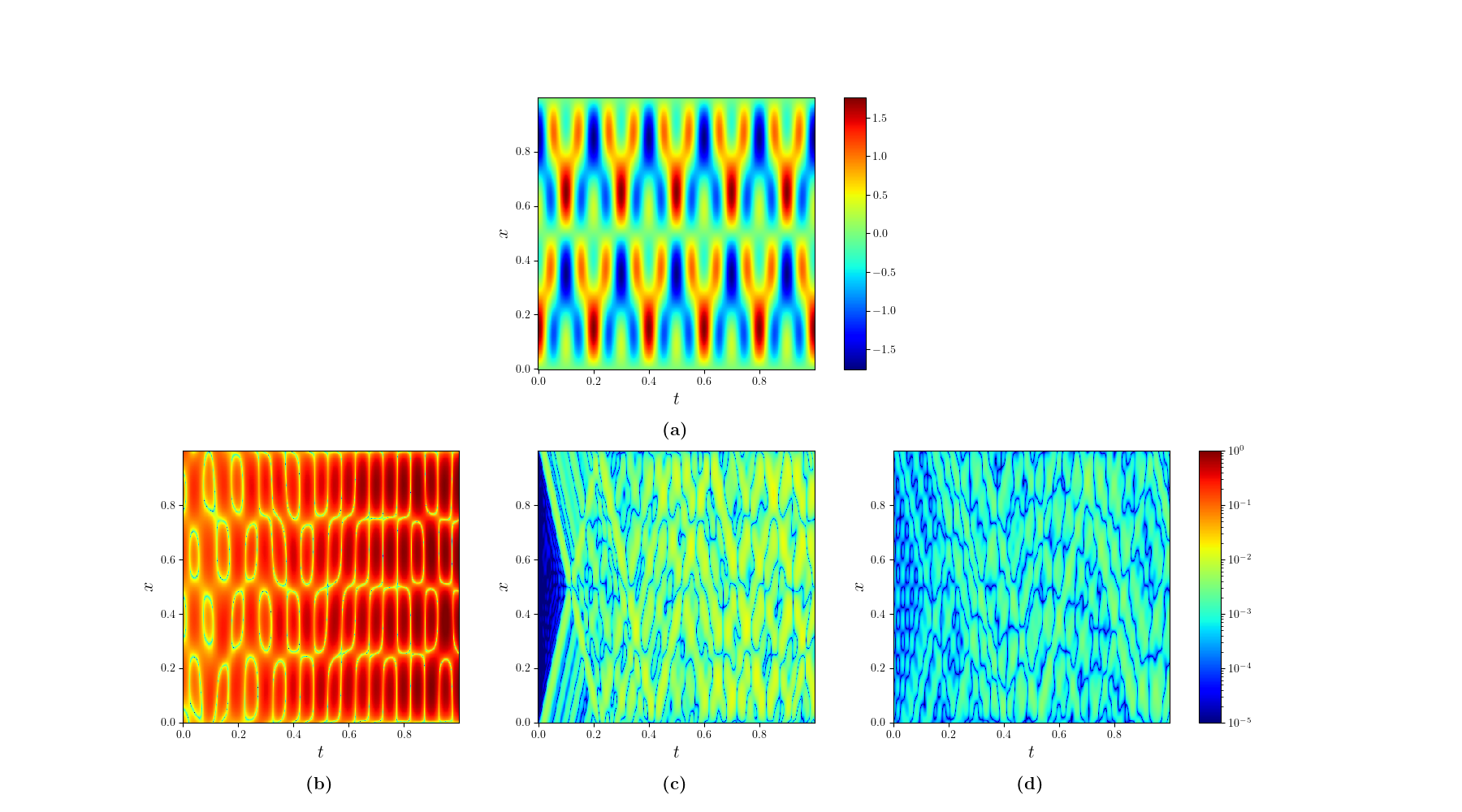}
  \caption{Wave equation. (a) The exact solution $u(x, t)$.
  (b) Step $It=0$, absolute error $|u - u_{\text{net},0}|$.
  (c) Step $It=1$, absolute error $|u - u_{\text{net},1}|$. 
  (d) Step $It=2$, absolute error $|u - u_{\text{net},2}|$.}
  \label{wa_err_all}
\end{figure}

\begin{figure}[H]
	\centering
  \includegraphics[width=0.8\linewidth]{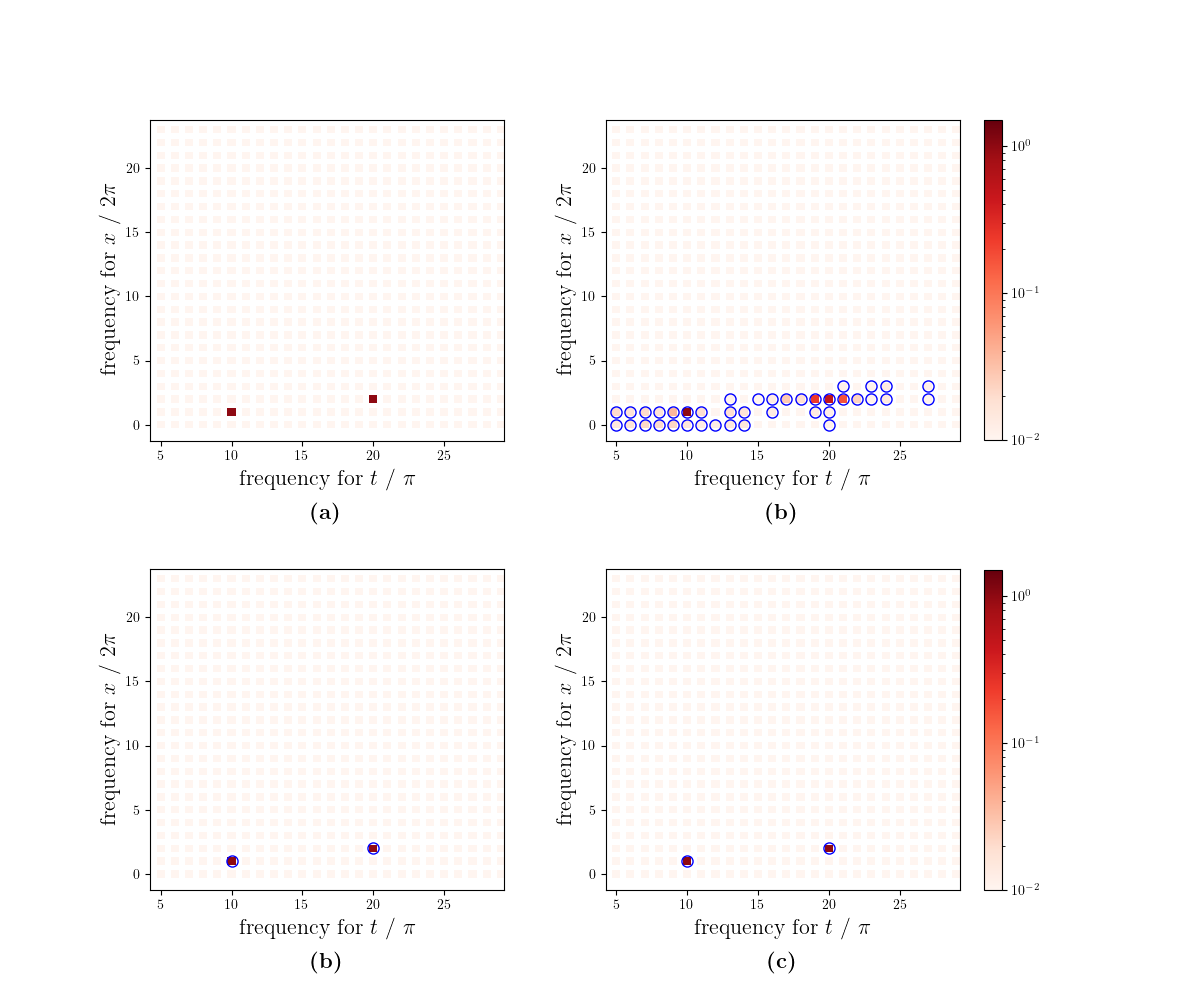}
  \caption{Wave equation.
  (a) Exact solution, distribution of Fourier coefficients.
  (b) Step $It=0$, distribution of Fourier coefficients $|\hat{u}_{\text{net},0,\bm{k}}|$.
  (c) Step $It=1$, distribution of Fourier coefficients $|\hat{u}_{\text{net},0,\bm{k}}|$.
  (d) Step $It=2$, distribution of Fourier coefficients $|\hat{u}_{\text{net},0,\bm{k}}|$. In Figures (b)-(d), a blue circle located at $\bm{k}$ means that the frequency $\bm{k}$ has been identified in $\mathbb{B}_{It}$.}
  \label{wa_se}
\end{figure}

 
We employ the frequency-adaptive MscaleDNNs with $I=4$ 
to solve the wave equation (\ref{wave_eq}). 
The parameter $\lambda$ is set to $0.01$.
Initially, the MscaleDNNs consist 6 sub networks with a multi-input 
frequency set $\mathbb{B}_0=\{2^0,\ldots, \ 2^5\}$, and the 
parameters for each sub network are $[1,100,100,100,1]$. 
The learning rate 
decays exponentially at a rate of 0.8 every 5000 
training iterations. 
At each adaptive step, we train the neural network for 100,000 epochs to 
get the solution $u_{\text{net},It}$. 
The numerical results for this simulation are reported 
in Figure \ref{wa_err_all} and Figure \ref{wa_se}. As the 
solution of equation (\ref{wave_eq}) involves high frequencies in time, 
the absolute error $|u - u_{\text{net},0}|$ shown 
in Figure \ref{wa_err_all}-(b) 
remains significantly large. After performing 
DFT on $u_{\text{net},0}$, 
the distribution of Fourier coefficients is 
displayed in Figure \ref{wa_se}-(b), which 
is similar to that of the exact solution reported in Figure \ref{wa_se}-(a). 

Based on the distribution of Fourier coefficients, we identify the feature set $\mathbb{B}_1$, represented by the blue circles in Figure \ref{wa_se}-(b). The MscaleDNNs are then adaptively adjusted according to Criterion A.
After training for 100,000 epochs, we obtain 
the neural network solution $u_{\text{net},1}$, and 
the absolute error $|u-u_{\text{net},1}|$ is depicted in 
Figure \ref{wa_err_all}-(c). This error is almost tow orders 
of magnitude smaller than $|u - u_{\text{net},0}|$, which 
demonstrates the effectiveness of frequency-adaptive 
MscaleDNNs. The simulation is stopped at 
$It=2$  as the stopping condition $\mathbb{B}_2=\mathbb{B}_3=\{(2\pi,10\pi),(4\pi,20\pi)\}$ is satisfied, as shown in Figure \ref{wa_se}-(c),(d). 
The final feature set $\mathbb{B}_2$ is consistent with the compact support of $\hat{u}$. 
The relative $L_2$ errors at all adaptive iteration are listed in Table \ref{table1}. 
The frequency-adaptive MscaleDNNs obtain an accuracy solution with relative error being 
1.899e-03, which is two orders of magnitude smaller than that of standard MscaleDNNs.

\subsection{Schr$\ddot{\text{o}}$dinger equation near the semi-classical limit}
\label{sc}
The final test case we considered is the following schr$\ddot{\text{o}}$dinger equation near the semi-classical limit:
\begin{equation}
  \begin{split}
    &\bm{\Psi}_{t} = \frac{\mathrm{i} \varepsilon}{2} \bm{\Psi}_{xx} - \frac{\mathrm{i}}{\varepsilon}\bm{V}(x) \bm{\Psi},  \quad (x,t)\in (0,\pi) \times (0,T],\\
    &\bm{\Psi}(x,0)=A(x)\exp(\mathrm{i}\phi(x)/\varepsilon), \quad x \in (0,\pi),
  \end{split}
  \label{sc_eq}
\end{equation}
where $\bm{\Psi}:=\bm{\Psi}(x,t) \in \mathbb{C}$ is a complex-valued function, $ \ A(x) \ \text{and} \ \phi(x)$ are given smooth 
functions. In equation (\ref{sc_eq}), $\varepsilon<1$ denotes the non-dimensional Planck's 
constant, and $\bm{V}(x)$ is the potential function. For this simulation, we consider
$V(x)=\frac{1}{2}x^2$, $\varepsilon=0.05,$ and a special initial value:
\begin{equation*}
	\bm{\Psi}(x,0)=\exp\biggl\{\frac{\mathrm{i}}{\varepsilon}\left[\frac{1}{2}\mathrm{i} (x - 1)^2 + 2(x - 1) + \frac{1}{4}\ln(\frac{1}{\pi \varepsilon})\right]\biggr\}. 
\end{equation*}
Periodic boundary condition is applied in the $x$ direction. 
To avoid complex computation, we decompose $\bm{\Psi}$ into the real 
part $\bm{\Psi}^{\text{re}}$ and the imaginary part $\bm{\Psi}^{\text{im}}$. 
Similarly, $\bm{\Psi}^{\text{re}}_{\text{net},It}$, $\bm{\Psi}^{\text{im}}_{\text{net},It}$ represent the 
two outputs of frequency-adaptive MscaleDNNs. The solution for (\ref{sc_eq}) 
obtained by frequency-adaptive MscaleDNNs is then given by
$\bm{\Psi}_{\text{net},It} = \bm{\Psi}^{\text{re}}_{\text{net},It}+\mathrm{i}\bm{\Psi}^{\text{im}}_{\text{net},It}$. 
The definition of the loss function for equation (\ref{sc_eq}) is similar to that of equation (\ref{wave_eq}). \par 

\begin{figure}[htbp]
	\centering
  \includegraphics[width=1.0\linewidth]{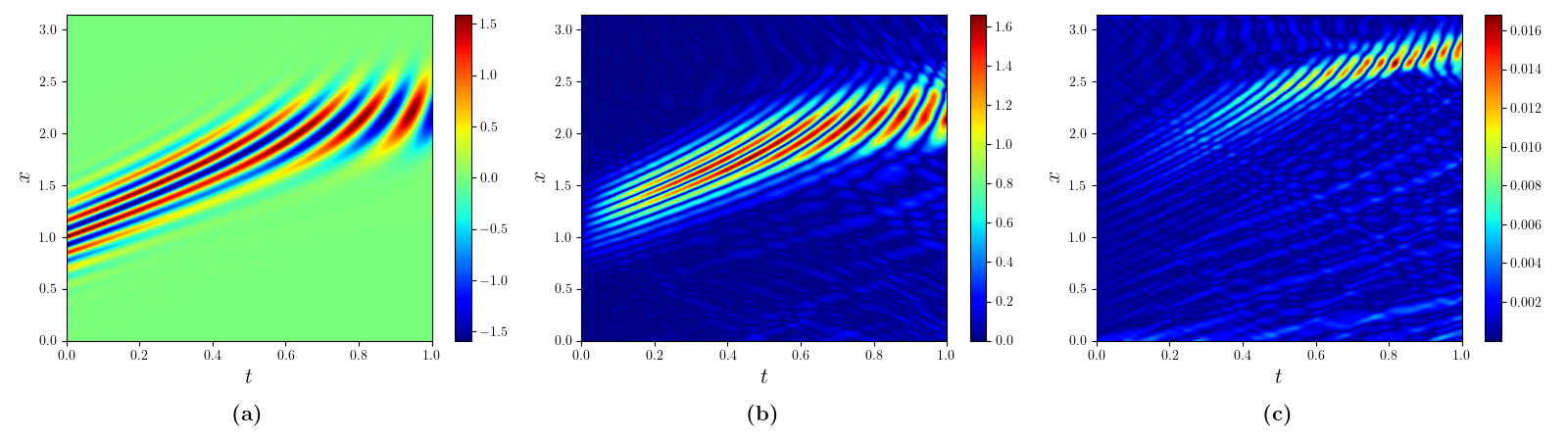}
   \includegraphics[width=1.0\linewidth]{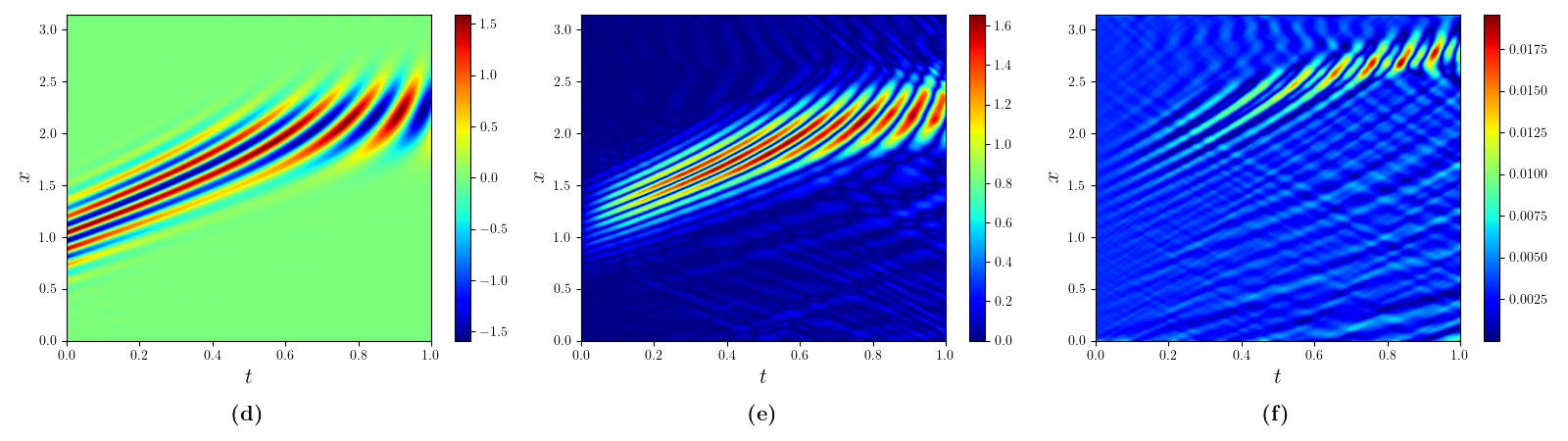}
  \caption{Schr$\ddot{\text{o}}$dinger equation near the semi-classical 
  limit. (a) Real part of exact solution $\bm{\Psi}^{\text{re}}$. 
  (b) Step $It=0$, absolute error $|\bm{\Psi}_{\text{net},0}^{\text{re}} - \bm{\Psi}^{\text{re}}|$. 
  (c) Step $It=4$, absolute error $|\bm{\Psi}_{\text{net},4}^{\text{re}} - \bm{\Psi}^{\text{re}}|$. (d) Imaginary part of exact solution $\bm{\Psi}^{\text{im}}$. 
  (e) Step $It=0$, absolute error $|\bm{\Psi}_{\text{net},0}^{\text{im}} - \bm{\Psi}^{\text{im}}|$. 
  (f) Step $It=4$, absolute error $|\bm{\Psi}_{\text{net},4}^{\text{im}} - \bm{\Psi}^{\text{im}}|$.}
  \label{sc_err_re}
\end{figure}

\begin{figure}
	\centering
  \includegraphics[width=1.0\linewidth]{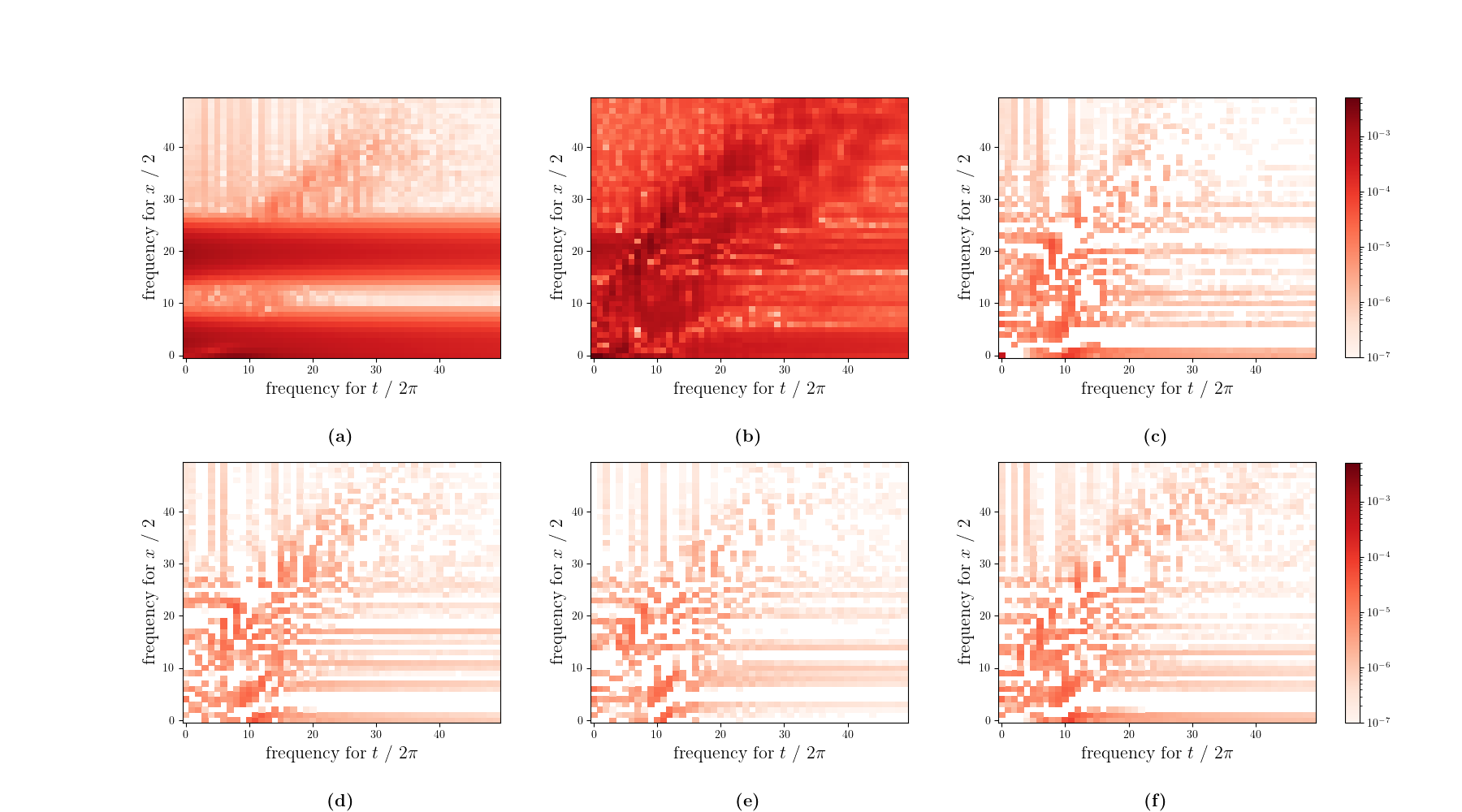}
  \caption{Schr$\ddot{\text{o}}$dinger equation near the semi-classical limit.
  (a) Distribution of Fourier coefficients $\hat{\bm{\Psi}}^{\text{re}}$.
  (b) Step $It=0$, distribution of $|\hat{\bm{\Psi}}^{\text{re}}_{\text{net},0,\bm{k}} - \hat{\bm{\Psi}}^{\text{re}}_{\bm{k}}|$.
  (c) Step $It=1$, distribution of $|\hat{\bm{\Psi}}^{\text{re}}_{\text{net},1,\bm{k}} - \hat{\bm{\Psi}}^{\text{re}}_{\bm{k}}|$.
  (d) Step $It=2$, distribution of $|\hat{\bm{\Psi}}^{\text{re}}_{\text{net},2,\bm{k}} - \hat{\bm{\Psi}}^{\text{re}}_{\bm{k}}|$.
  (e) Step $It=3$, distribution of $|\hat{\bm{\Psi}}^{\text{re}}_{\text{net},3,\bm{k}} - \hat{\bm{\Psi}}^{\text{re}}_{\bm{k}}|$.
  (f) Step $It=4$, distribution of $|\hat{\bm{\Psi}}^{\text{re}}_{\text{net},4,\bm{k}} - \hat{\bm{\Psi}}^{\text{re}}_{\bm{k}}|$.}
  \label{sc_fre}
\end{figure}

\begin{figure}
	\centering
  \includegraphics[width=1.0\linewidth]{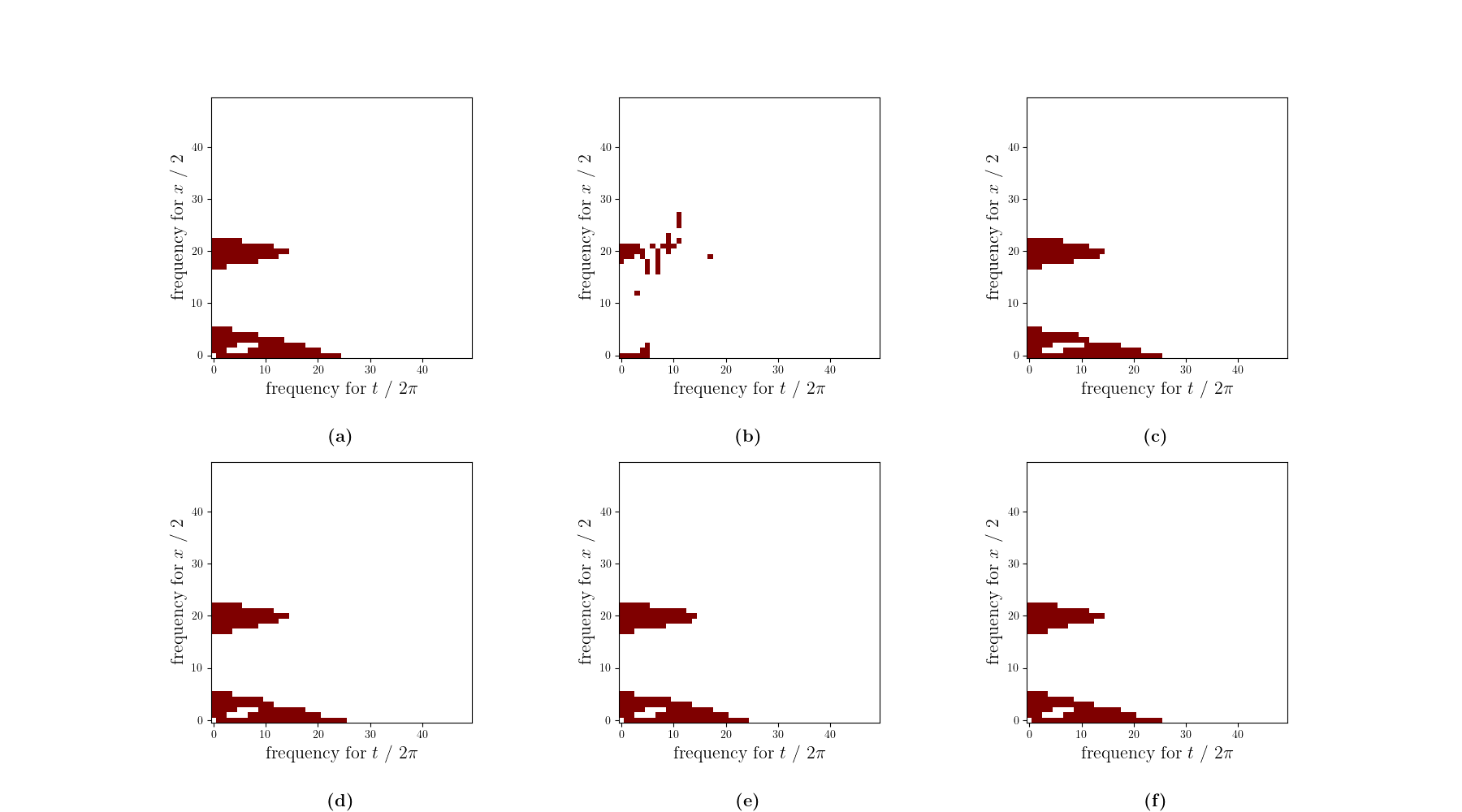}
  \caption{Schr$\ddot{\text{o}}$dinger equation near the semi-classical limit.
  (a) The frequency $\bm{k}$ in dark part satisfies $|\hat{\bm\Psi}^\text{re}_{\bm{k}}| > 0.1 \max_{\bm{k} \in \mathbb{B}} |\hat{\bm\Psi}^\text{re}_{\bm{k}}|$.
  (b) Step $It=0$, the frequency $\bm{k}$ in dark part belongs to $\mathbb{B}_1$.
  (c) Step $It=1$, the frequency $\bm{k}$ in dark part belongs to $\mathbb{B}_2$.
  (d) Step $It=2$, the frequency $\bm{k}$ in dark part belongs to $\mathbb{B}_3$.
  (e) Step $It=3$, the frequency $\bm{k}$ in dark part belongs to $\mathbb{B}_4$.
  (f) Step $It=4$, the frequency $\bm{k}$ in dark part belongs to $\mathbb{B}_5$. }
  \label{sc_se}
\end{figure}

We employ the frequency-adaptive MscaleDNNs with $I=4$ to solve the equation (\ref{sc_eq}). 
Initially, the networks have 6 sub networks with the multiple inputs set $\mathbb{B}_0= \{2^0, \ldots,\ 2^5\}$.
The parameters for each sub network are $[2,200,200,200,2]$. The learning rate decays exponentially at a rate of 0.9 every 1000 training iterations. 
Figure \ref{sc_err_re} shows the absolute errors of 
$|\bm{\Psi}_{\text{net},0}^{\text{re}}-\bm{\Psi}^{\text{re}}|$
and $|\bm{\Psi}_{\text{net},0}^{\text{im}}-\bm{\Psi}^{\text{im}}|$, which are larger than 1 in many points.
The absolute error of the Fourier coefficients for 
$\bm{\Psi}_{\text{net},0}^{\text{re}}$ and $\bm{\Psi}^{\text{re}}$ is 
displayed in Figure \ref{sc_fre}-(b), where the error remains relatively large. By setting $\lambda=0.1$, we 
obtain the frequency feature set $\mathbb{B}_1$, as shown in Figure \ref{sc_se}-(b). Although the error is still significant, a comparison between Figures \ref{sc_se}-(a) and (b) reveals that the first high-contribution frequency region near $k_x=20$ for $\bm{\Psi}$ is captured within $\mathbb{B}_1$.  
Consequently, we use the frequency feature set $\mathbb{B}_1$ 
to adaptively adjust MscaleDNNs at $It=1$ following Criterion B. After training, we 
get a new solution $\bm{\Psi}_{\text{net},1}$ with the relative $L_2$ error 
reduced to 1.280e-02, as shown in Table \ref{table1}.

The evolution of $\mathbb{B}_{It}$ is displayed in Figure \ref{sc_se}. At adaptive step $It=2$, another high-contribution frequency region near $k_x=0$ for $\bm{\Psi}$ is successfully captured within $\mathbb{B}_2$. The feature set $\mathbb{B}_{It}$ is then fine-tuned through subsequent adaptive iterations,  ultimately aligning well with the reference solution by $It=4$. The distributions of the Fourier coefficients for 
$\bm{\Psi}^{\text{re}}_{\text{net},It}$ at $It=1,2,3$, and 4 are exhibited in Figure \ref{sc_fre}, 
which demonstrates good agreement with that of $\bm{\Psi}^{\text{re}}$.
The simulation is stopped at $It=4$, and 
the absolute error for $\bm{\Psi}_{\text{net},4}$ is given in Figure \ref{sc_err_re}-(c) (real part) 
and Figure \ref{sc_err_re}-(f) (imaginary part), both of which demonstrate significant improvement compared to  $\bm{\Psi}_{\text{net},0}$. The relative $L_2$ error at each iteration is reported in 
Table \ref{table1}. These results  confirm 
the effectiveness and accuracy of the frequency-adaptive MscaleDNNs.\par

\section{Conclusion}
\label{co}
In this study, we established a fitting error bound for MscaleDNNs and introduced frequency-adaptive MscaleDNNs with a robust feature embedding. These method present new techniques for capturing frequency features from the solution and embedding them into MscaleDNNs. Unlike traditional MscaleDNNs, which pre-define a frequency range, our approach effectively handles higher frequencies and more pronounced multi-scale behaviors through adaptive frequency adjustment.
Positive results have been observed in solving the Poisson equation, heat equation, wave equation, and Schr$\ddot{\text{o}}$dinger equation 
near the semi-classical limit. 
However, we acknowledge that the current adaptive frequency method is relatively basic and has certain limitations. One notable drawback is the need to retrain new networks for each adaptation, which leads to higher computational costs. There is potential for improvement by exploring methods to project the original solution into the newly designed network to facilitate training. Additionally, for highly complex problems, the initial MscaleDNNs may fail to capture the frequency features of the exact solution, making subsequent frequency adaptation ineffective. Addressing these challenges presents a promising direction for future research.

\section*{Acknowledgement}
\nocite{*}

\bibliographystyle{elsarticle-num} 

\bibliography{ref}	

\begin{thebibliography}{10}
\expandafter\ifx\csname url\endcsname\relax
  \def\url#1{\texttt{#1}}\fi
\expandafter\ifx\csname urlprefix\endcsname\relax\def\urlprefix{URL }\fi
\expandafter\ifx\csname href\endcsname\relax
  \def\href#1#2{#2} \def\path#1{#1}\fi

\bibitem{brown1990statistical}
P.~F. Brown, J.~Cocke, S.~A. Della~Pietra, V.~J. Della~Pietra, F.~Jelinek, J.~Lafferty, R.~L. Mercer, P.~S. Roossin, A statistical approach to machine translation, Computational linguistics 16~(2) (1990) 79--85.

\bibitem{vaswani2017attention}
A.~Vaswani, Attention is all you need, arXiv preprint arXiv:1706.03762 (2017).

\bibitem{devlin2018bert}
J.~Devlin, Bert: Pre-training of deep bidirectional transformers for language understanding, arXiv preprint arXiv:1810.04805 (2018).

\bibitem{brown2020language}
T.~B. Brown, Language models are few-shot learners, arXiv preprint ArXiv:2005.14165 (2020).

\bibitem{han2017deep}
J.~Han, A.~Jentzen, et~al., Deep learning-based numerical methods for high-dimensional parabolic partial differential equations and backward stochastic differential equations, Communications in mathematics and statistics 5~(4) (2017) 349--380.

\bibitem{yu2018deep}
B.~Yu, et~al., The deep ritz method: a deep learning-based numerical algorithm for solving variational problems, Communications in Mathematics and Statistics 6~(1) (2018) 1--12.

\bibitem{zang2020weak}
Y.~Zang, G.~Bao, X.~Ye, H.~Zhou, Weak adversarial networks for high-dimensional partial differential equations, Journal of Computational Physics 411 (2020) 109409.

\bibitem{tran2019dnn}
T.~Tran, A.~Hamilton, M.~B. McKay, B.~Quiring, P.~S. Vassilevski, Dnn approximation of nonlinear finite element equations, arXiv preprint arXiv:1911.05240 (2019).

\bibitem{han2018solving}
J.~Han, A.~Jentzen, W.~E, Solving high-dimensional partial differential equations using deep learning, Proceedings of the National Academy of Sciences 115~(34) (2018) 8505--8510.

\bibitem{han2017deep2}
J.~Han, L.~Zhang, R.~Car, et~al., Deep potential: A general representation of a many-body potential energy surface, arXiv preprint arXiv:1707.01478 (2017).

\bibitem{he2018relu}
J.~He, L.~Li, J.~Xu, C.~Zheng, Relu deep neural networks and linear finite elements, arXiv preprint arXiv:1807.03973 (2018).

\bibitem{liao2019deep}
Y.~Liao, P.~Ming, Deep nitsche method: Deep ritz method with essential boundary conditions, arXiv preprint arXiv:1912.01309 (2019).

\bibitem{raissi2019physics}
M.~Raissi, P.~Perdikaris, G.~E. Karniadakis, Physics-informed neural networks: A deep learning framework for solving forward and inverse problems involving nonlinear partial differential equations, Journal of Computational physics 378 (2019) 686--707.

\bibitem{strofer2019data}
C.~M. Strofer, J.-L. Wu, H.~Xiao, E.~Paterson, Data-driven, physics-based feature extraction from fluid flow fields using convolutional neural networks, Communications in Computational Physics 25~(3) (2019) 625--650.

\bibitem{wang2020mesh}
Z.~Wang, Z.~Zhang, A mesh-free method for interface problems using the deep learning approach, Journal of Computational Physics 400 (2020) 108963.

\bibitem{rahaman2019spectral}
N.~Rahaman, A.~Baratin, D.~Arpit, F.~Draxler, M.~Lin, F.~Hamprecht, Y.~Bengio, A.~Courville, On the spectral bias of neural networks, in: International Conference on Machine Learning, PMLR, 2019, pp. 5301--5310.

\bibitem{xu2019frequency}
Z.-Q.~J. Xu, Y.~Zhang, T.~Luo, Y.~Xiao, Z.~Ma, Frequency principle: Fourier analysis sheds light on deep neural networks, arXiv preprint arXiv:1901.06523 (2019).

\bibitem{xu2019training}
Z.-Q.~J. Xu, Y.~Zhang, Y.~Xiao, Training behavior of deep neural network in frequency domain, in: Neural Information Processing: 26th International Conference, ICONIP 2019, Sydney, NSW, Australia, December 12--15, 2019, Proceedings, Part I 26, Springer, 2019, pp. 264--274.

\bibitem{zhang2019explicitizing}
Y.~Zhang, Z.-Q.~J. Xu, T.~Luo, Z.~Ma, Explicitizing an implicit bias of the frequency principle in two-layer neural networks, arXiv preprint arXiv:1905.10264 (2019).

\bibitem{xu2018understanding}
Z.~J. Xu, Understanding training and generalization in deep learning by fourier analysis, arXiv preprint arXiv:1808.04295 (2018).

\bibitem{langer2021approximating}
S.~Langer, Approximating smooth functions by deep neural networks with sigmoid activation function, Journal of Multivariate Analysis 182 (2021) 104696.

\bibitem{lu2021deep}
J.~Lu, Z.~Shen, H.~Yang, S.~Zhang, Deep network approximation for smooth functions, SIAM Journal on Mathematical Analysis 53~(5) (2021) 5465--5506.

\bibitem{shen2019deep}
Z.~Shen, H.~Yang, S.~Zhang, Deep network approximation characterized by number of neurons. arxiv e-prints, page, arXiv preprint arXiv:1906.05497 (2019).

\bibitem{deng2018learning}
M.~Deng, S.~Li, G.~Barbastathis, Learning to synthesize: splitting and recombining low and high spatial frequencies for image recovery, arXiv preprint arXiv:1811.07945 (2018).

\bibitem{pan2018learning}
J.~Pan, S.~Liu, D.~Sun, J.~Zhang, Y.~Liu, J.~Ren, Z.~Li, J.~Tang, H.~Lu, Y.-W. Tai, et~al., Learning dual convolutional neural networks for low-level vision, in: Proceedings of the IEEE conference on computer vision and pattern recognition, 2018, pp. 3070--3079.

\bibitem{wu2020multigrid}
C.-Y. Wu, R.~Girshick, K.~He, C.~Feichtenhofer, P.~Krahenbuhl, A multigrid method for efficiently training video models, in: Proceedings of the IEEE/CVF Conference on Computer Vision and Pattern Recognition, 2020, pp. 153--162.

\bibitem{biland2019frequency}
S.~Biland, V.~C. Azevedo, B.~Kim, B.~Solenthaler, Frequency-aware reconstruction of fluid simulations with generative networks, arXiv preprint arXiv:1912.08776 (2019).

\bibitem{cai2020phase}
W.~Cai, X.~Li, L.~Liu, A phase shift deep neural network for high frequency approximation and wave problems, SIAM Journal on Scientific Computing 42~(5) (2020) A3285--A3312.

\bibitem{liu2020multi}
Z.~Liu, W.~Cai, Z.-Q.~J. Xu, Multi-scale deep neural network (mscalednn) for solving poisson-boltzmann equation in complex domains, arXiv preprint arXiv:2007.11207 (2020).

\bibitem{jacot2018neural}
A.~Jacot, F.~Gabriel, C.~Hongler, Neural tangent kernel: Convergence and generalization in neural networks, Advances in neural information processing systems 31 (2018).

\bibitem{tancik2020fourier}
M.~Tancik, P.~Srinivasan, B.~Mildenhall, S.~Fridovich-Keil, N.~Raghavan, U.~Singhal, R.~Ramamoorthi, J.~Barron, R.~Ng, Fourier features let networks learn high frequency functions in low dimensional domains, Advances in Neural Information Processing Systems 33 (2020) 7537--7547.

\bibitem{zhong2019reconstructing}
E.~D. Zhong, T.~Bepler, J.~H. Davis, B.~Berger, Reconstructing continuous distributions of 3d protein structure from cryo-em images, arXiv preprint arXiv:1909.05215 (2019).

\bibitem{rahimi2007random}
A.~Rahimi, B.~Recht, Random features for large-scale kernel machines, Advances in neural information processing systems 20 (2007).

\bibitem{wang2022and}
S.~Wang, X.~Yu, P.~Perdikaris, When and why pinns fail to train: A neural tangent kernel perspective, Journal of Computational Physics 449 (2022) 110768.

\bibitem{wang2021eigenvector}
S.~Wang, H.~Wang, P.~Perdikaris, On the eigenvector bias of fourier feature networks: From regression to solving multi-scale pdes with physics-informed neural networks, Computer Methods in Applied Mechanics and Engineering 384 (2021) 113938.

\bibitem{wang2022spectral}
B.~Wang, H.~Yuan, L.~Liu, W.~Zhang, W.~Cai, On spectral bias reduction of multi-scale neural networks for regression problems, arXiv preprint arXiv:2212.03416 (2022).

\bibitem{wu2023comprehensive}
C.~Wu, M.~Zhu, Q.~Tan, Y.~Kartha, L.~Lu, A comprehensive study of non-adaptive and residual-based adaptive sampling for physics-informed neural networks, Computer Methods in Applied Mechanics and Engineering 403 (2023) 115671.

\bibitem{gao2023failure}
Z.~Gao, L.~Yan, T.~Zhou, Failure-informed adaptive sampling for pinns, SIAM Journal on Scientific Computing 45~(4) (2023) A1971--A1994.

\bibitem{tang2023pinns}
K.~Tang, X.~Wan, C.~Yang, Das-pinns: A deep adaptive sampling method for solving high-dimensional partial differential equations, Journal of Computational Physics 476 (2023) 111868.

\bibitem{glorot2010understanding}
X.~Glorot, Y.~Bengio, Understanding the difficulty of training deep feedforward neural networks, in: Proceedings of the thirteenth international conference on artificial intelligence and statistics, JMLR Workshop and Conference Proceedings, 2010, pp. 249--256.

\bibitem{kingma2014adam}
D.~P. Kingma, J.~Ba, Adam: A method for stochastic optimization, arXiv preprint arXiv:1412.6980 (2014).

\bibitem{li2023subspace}
X.-A. Li, Z.-Q.~J. Xu, L.~Zhang, Subspace decomposition based dnn algorithm for elliptic type multi-scale pdes, Journal of Computational Physics 488 (2023) 112242.

\bibitem{lagaris1998artificial}
I.~E. Lagaris, A.~Likas, D.~I. Fotiadis, Artificial neural networks for solving ordinary and partial differential equations, IEEE transactions on neural networks 9~(5) (1998) 987--1000.

\bibitem{daw2022mitigating}
A.~Daw, J.~Bu, S.~Wang, P.~Perdikaris, A.~Karpatne, Mitigating propagation failures in pinns using evolutionary sampling (2022).

\bibitem{jagtap2020adaptive}
A.~D. Jagtap, K.~Kawaguchi, G.~E. Karniadakis, Adaptive activation functions accelerate convergence in deep and physics-informed neural networks, Journal of Computational Physics 404 (2020) 109136.

\bibitem{basri2020frequency}
R.~Basri, M.~Galun, A.~Geifman, D.~Jacobs, Y.~Kasten, S.~Kritchman, Frequency bias in neural networks for input of non-uniform density, in: International Conference on Machine Learning, PMLR, 2020, pp. 685--694.

\bibitem{heller1975time}
E.~J. Heller, Time-dependent approach to semiclassical dynamics, The Journal of Chemical Physics 62~(4) (1975) 1544--1555.

\bibitem{howell2016principles}
K.~B. Howell, Principles of Fourier analysis, CRC Press, 2016.

\end{thebibliography}


\end{document}